\documentclass[11pt]{article}

\usepackage[final]{acl}
\usepackage{times}
\usepackage{latexsym}
\usepackage[T1]{fontenc}
\usepackage[utf8]{inputenc}
\usepackage{microtype}

\usepackage{inconsolata}
\usepackage{graphicx}
\usepackage{float}
\usepackage{placeins}
\usepackage{flushend}
\usepackage{soul}

\usepackage{amsmath}
\usepackage{amssymb}
\usepackage{booktabs}
\usepackage{multirow}
\usepackage{xcolor}
\usepackage{algorithm}
\usepackage{algorithmic}
\usepackage{enumitem}
\usepackage{tcolorbox}
\usepackage{listings}
\usepackage{cleveref}

\tcbuselibrary{breakable}
\newtcolorbox{promptbox}[1][]{
  colback=gray!5,
  colframe=black!75,
  fonttitle=\bfseries\small,
  boxrule=0.5pt,
  arc=2pt,
  left=6pt,
  right=6pt,
  top=4pt,
  bottom=4pt,
  breakable,
  #1
}

\usepackage{pifont}
\newcommand{\yes}{\textcolor{green!60!black}{\ding{51}}}
\newcommand{\no}{\textcolor{red!70!black}{\ding{55}}}

\title{Formula-One Prompting: A Composable Equation-First Prefix \\ for Applied Mathematics}

\author{
 \textbf{Natapong Nitarach}, 
 \textbf{Pittawat Taveekitworachai}\thanks{\ \ Equal advising.},
 \textbf{Kunat Pipatanakul}\footnotemark[1]
\\
\\
 SCB DataX, SCBX Group
}

\begin{document}
\maketitle

\begin{abstract}
This paper introduces \textbf{Formula Prompting} (FP) and \textbf{Formula-One Prompting} (F-1), two single-call methods that elicit governing equations before solving applied-math problems. Chain-of-Thought (CoT) and Program-of-Thought (PoT) prompting improve mathematical reasoning by eliciting reasoning traces or code-like structures learned during pretraining. This suggests a diagnostic question: which useful pretraining patterns remain under-elicited? Using \textsc{infini-gram-mini}, we scan 81.7 trillion pretraining tokens and find that, in curated corpora such as DataComp-LM, equation-centered language appears 121$\times$ more often than code and 3.79$\times$ more often than step-by-step narration, yet standard prompting methods do not explicitly elicit equation formulation. FP asks the model to formalize a problem's governing equations before solving; F-1 extends FP with a composable Phase 2 that selects Direct, CoT, or PoT-style solving in the same call. Across five reasoning models and four applied-math benchmarks (finance, physics, cryptography, competition math), F-1 outperforms CoT by 5.76\,pp and PoT by 8.42\,pp on average, with the largest gain of 13.30\,pp on FinanceMath, while topping the accuracy-token efficiency frontier at only 68 prompt tokens of overhead. Ablations show that the equation-formalization prefix, rather than the Phase-2 strategy menu, drives most of the gains: adding CoT or PoT after the prefix provides no further improvement, and 73.3\% of remaining errors occur despite a correct Phase-1 equation.
\end{abstract}

\begin{figure}[!t]
  \centering
  \includegraphics[width=\columnwidth]{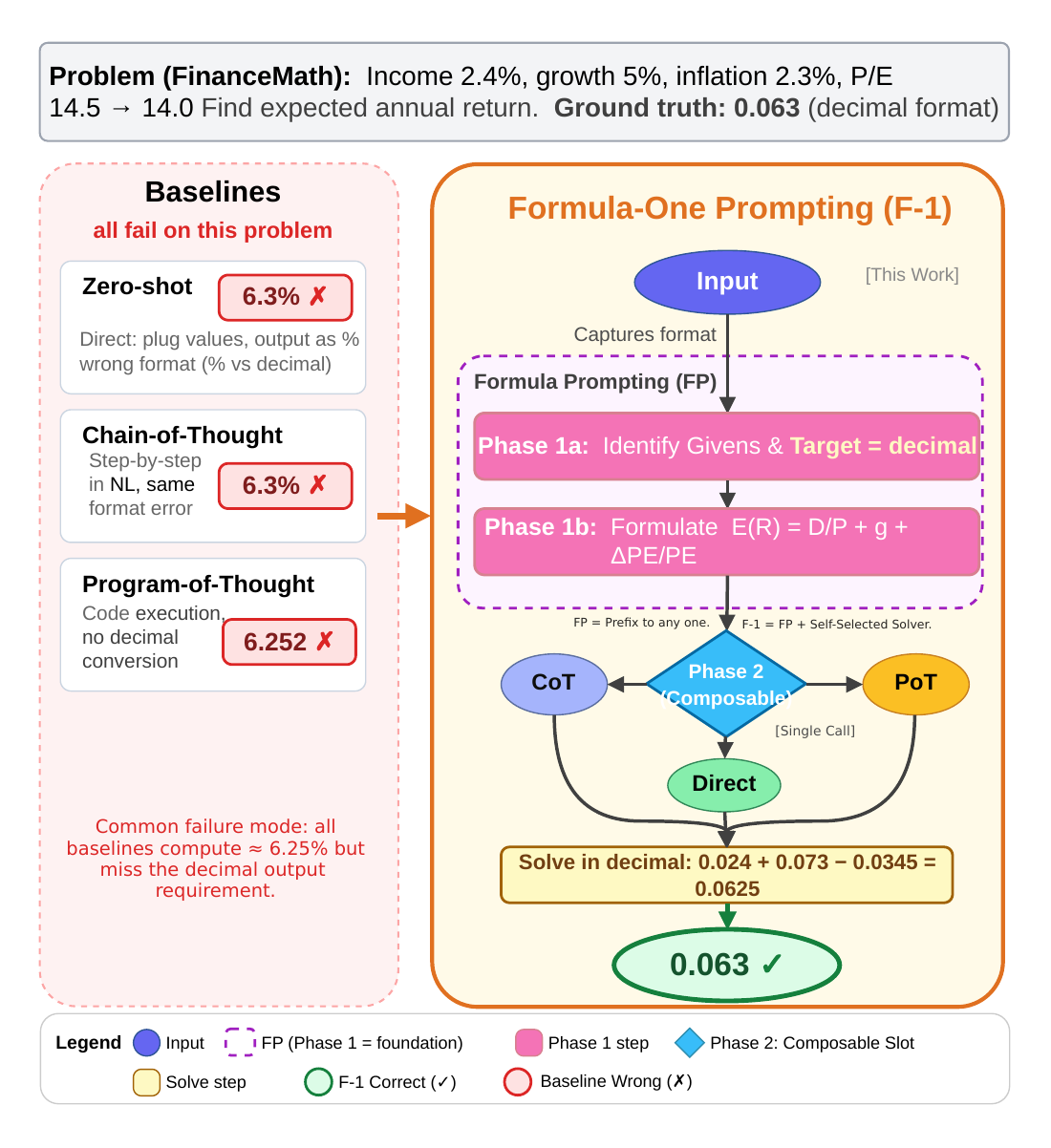}
  \vspace{-0.5em}
  \caption{F-1 compared to other prompting methods on a FinanceMath problem (about Grinold-Kroner; ground truth is $0.063$). Baselines compute $\approx$6.25\% but miss the decimal-format requirement; F-1's Phase 1 captures both equation and format, and the Phase 2 chooses any solving strategy.}
  \label{fig:overview}
\end{figure}

\section{Introduction}

Chain-of-Thought (CoT) \citep{wei2022chain,kojima2022large} and Program-of-Thought (PoT) \citep{chen2023program} show that prompting can improve mathematical reasoning by eliciting formats present in pretraining \citep{prystawski2023why,mccoy2024embers}, including step-by-step explanations and code. We later quantify this prevalence directly (\Cref{sec:infinigram}). Applied mathematics, e.g., problems in finance, physics, and cryptography, often depend on domain-specific \textit{equations}, including compound interest formulas, Newton's laws, and security game definitions \citep{wang2024scibench,xu2025ugphysics}. Large math corpora also contain substantial LaTeX and equation-heavy content \citep{lewkowycz2022minerva,paster2023openwebmath,shao2024deepseekmath}. Therefore, we investigate the effectiveness of \emph{explicitly} eliciting a model to formulate equations before solving a problem. This contrasts with existing prompting approaches such as CoT and PoT, which may generate equations implicitly during the solving process, but do not consistently require them.

To measure this opportunity, we run an infini-gram-mini scan over 81.7T tokens (\Cref{sec:infinigram}). In DCLM \citep{li2025datacomplmsearchgenerationtraining}, equation-language, including LaTeX typography patterns, occurs 121$\times$ more often than code and 3.79$\times$ more often than step-by-step narration. Following evidence that pretraining frequency predicts downstream task performance \citep{razeghi2022impact,prystawski2023why,mccoy2024embers}, we use this prevalence as a diagnostic signal: equation-language patterns may be an \emph{under-used promptable pattern}. Existing prompting methods target other forms of mathematical reasoning--CoT elicits natural-language reasoning, PoT elicits code generation, and Plan-and-Solve \citep{wang2023planandsolve} elicits planning--but they do not explicitly ask the model to formulate the governing equations central to problem solving.

To understand the effectiveness of explicit formulation-first instruction, which we call \textbf{Formula Prompting (FP)}, we compare FP against a zero-shot baseline and run four analyses to study what the prefix contributes: per-problem flips against a fixed CoT solver (\Cref{sec:per_problem}), output-pattern composability across solvers (\Cref{sec:blending}), variant accuracy at matched reasoning budget (\Cref{sec:variant_accuracy}), and a failure-cause classification showing that 73.3\% of remaining failures have a correct Phase-1 equation but fail in the downstream solver (\Cref{sec:error_classification}). Motivated by these analyses, we extend FP into \textbf{Formula-One Prompting (F-1)} (\Cref{fig:overview}) with a composable second phase that selects Direct, CoT, or PoT solving within a single inference call, preserving equation formalization as the shared intervention. F-1 outperforms CoT by 5.76\,pp and PoT by 8.42\,pp on average. While Formulate-and-Solve prompting \citep{kao2024formulate} also formulates equations, F-1 separates equation formulation from the composable selection of a solving strategy (\Cref{tab:comparison}). Our contributions are:

\begin{itemize}[leftmargin=*]
\setlength\itemsep{-0.1em}
    \item \textbf{Formula Prompting (FP)} A novel equation-first prompting approach. Our experiments show that explicitly eliciting equation formulation before solving is effective compared with a zero-shot baseline in most included model-benchmark combinations (\Cref{sec:fp}).
    \item \textbf{Formula-One Prompting (F-1)} A composable extension of FP that adds a second phase selecting Direct, CoT, or PoT solving within a single call. Experiments show that F-1 outperforms CoT and PoT prompting in most cases and achieves higher token efficiency than baselines (\Cref{sec:method}).
    \item \textbf{Pretraining-prevalence diagnostic} An infini-gram-mini scan over 81.7T pretraining tokens. We measure the relative prevalence of equation-language, code, and step-by-step narration patterns, supporting the motivation for explicit equation elicitation (\Cref{sec:infinigram}).
\end{itemize}

\section{Related Work}
\label{sec:related}

\subsection{Single-Call Prompt Engineering}
\textbf{Chain-of-Thought} \citep{wei2022chain,kojima2022large} improves LLM reasoning by eliciting step-by-step natural-language explanations, but its arithmetic and derivation steps can accumulate errors \citep{chen2023program,sprague2024cot}. \textbf{Program-of-Thought} \citep{chen2023program} reduces computation errors through executable code, but is most useful when problems are programmatically expressible; CoT and PoT have been shown to be complementary across reasoning tasks \citep{liu2023xot}. \textbf{Plan-and-Solve} \citep{wang2023planandsolve} adds explicit planning, but structures the execution process rather than the problem representation. None of these methods explicitly formalize governing equations before solving. For applied-math domains where governing equations are identifiable, such as finance, physics, and cryptography \citep{wang2024scibench,xu2025ugphysics}, this leaves the symbolic structure implicit during reasoning. For tasks without such structure, explicit formalization is not appropriate; our negative-control experiment (\Cref{sec:negative_control}) shows it can in fact hurt. 

Beyond single-call prompting, search and routing methods such as \textbf{ToT} \citep{yao2023tree}, \textbf{GoT} \citep{besta2024got}, \textbf{XoT} \citep{liu2023xot}, and \textbf{Adaptive-Solver} \citep{zhou2024adaptivesolver} explore multiple reasoning paths or route each problem to a chosen solver, at substantially higher computational cost (3 to 100+ model invocations per problem). F-1 targets the single-call regime and is complementary to these search-based approaches (\Cref{tab:comparison}).

Prior work explores equation formalization: MathPrompter \citep{imani2023mathprompter} uses generic algebraic templates, \citet{heyueya2023solving} rely on multi-call SymPy solving, and Formulate-and-Solve (F\&S) \citep{kao2024formulate} prompts equation-first solutions for algebra word problems. F-1 differs from this line of work along three axes. \textbf{(1) Single-call, no external tools.} F-1 needs no symbolic solver or external router (unlike \citealp{heyueya2023solving}). \textbf{(2) Composable Phase 2.} F-1 offers a Direct, CoT, or PoT menu within the same call (unlike F\&S's fixed algebraic solve). \textbf{(3) Applied-math focus.} 
Prior equation-based methods target generic algebra; F-1 targets applied-math domains such as finance, physics, and cryptography, whose governing equations are domain-canonical (e.g., compound interest, Newton's laws, security games) rather than ad-hoc per-problem expressions. This makes formalization a structural rather than incidental intervention. \Cref{tab:comparison} summarizes design differences; \Cref{tab:fs_head_to_head} reports a head-to-head accuracy comparison with F\&S in the main body, with full per-problem cross-tab in \Cref{app:fs_complementarity}.

\section{Pretraining Pattern Diagnostic}
\label{sec:infinigram}

\paragraph{Objective.} We test whether the empirical effectiveness of equation-first prompting is partly predicted by the prevalence of equation-language in pretraining, consistent with prior findings that pretraining frequency correlates with downstream task performance \citep{razeghi2022impact,prystawski2023why,mccoy2024embers}.

\paragraph{Method.} We query the Infini-gram-mini API \citep{xu-etal-2025-infini} for n-gram occurrences of method-associated patterns across eight pretraining corpora ($\sim$81.7T tokens): DCLM-baseline (16.7T curated tokens) and seven Common Crawl snapshots. For each method (CoT, PoT, FP) we instantiate two pattern classes: \textbf{prompt phrases} taken verbatim from its prompt template (step-by-step language for CoT, code instructions for PoT, equation-formalization language plus LaTeX environments for FP), and \textbf{solving vocabulary} the method tends to produce in outputs (math notation including LaTeX for FP, code imports for PoT). Per-pattern occurrence ratios on actual model outputs validate method-specificity: associated patterns appear at 10$\times$ or higher frequency in their own outputs than in others'. Full pattern lists and validation in \Cref{app:infini-gram}.

\paragraph{Hypothesis.} If prevalence predicts elicitability, FP-associated patterns should be at least as prevalent as CoT-associated patterns in curated corpora, and substantially more prevalent than PoT-associated patterns. The PoT gap would predict FP's largest lift over PoT in particular.

\paragraph{Results.} On DCLM (curated), FP-associated patterns occur \textbf{121$\times$} more often than PoT-associated patterns (69.6M vs.\ 0.6M occurrences) and \textbf{3.79$\times$} more often than CoT-associated patterns (69.6M vs.\ 18.4M). The single LaTeX command \verb|\frac{| alone (23.9M) exceeds all PoT patterns combined. The FP/PoT ratio remains 19--61$\times$ across the seven raw Common Crawl snapshots. To rule out that a few equation-heavy documents drive the count, we re-tabulated by unique source document rather than raw occurrences (\Cref{app:doc_norm}): each method averages $\approx$1.0 occurrence per containing document, so the ranking is not a concentration artifact. See \Cref{fig:eq-vs-code} in \Cref{app:infini-gram} for the full breakdown.

\paragraph{Interpretation and scope.} The diagnostic provides \emph{correlational} evidence: pattern abundance in pretraining is consistent with FP's empirical lift reported in later sections, but the mechanism behind this lift is out of scope. We use the analysis to motivate the design of FP and to bound its applicable domain; mechanistic claims about how models represent equations or why equation-first prompting changes their behavior are left for future work.

\section{Experimental Setup}
\label{sec:experiments}

All experiments in this paper follow the setup described here, unless explicitly stated otherwise.

\paragraph{Datasets.} We evaluate all prompting approaches on four applied-mathematics benchmarks where formulation-first reasoning may help, spanning physics problems that invoke Newton's laws or thermodynamic equations, finance problems requiring compound-interest or pricing models, and cryptography problems requiring security-game formalizations. The included benchmarks are listed in \Cref{tab:datasets}.

\begin{table}[H]
\centering
\small
\resizebox{\columnwidth}{!}{%
\begin{tabular}{@{}llrl@{}}
\toprule
\textbf{Benchmark} & \textbf{Domain} & \textbf{N} & \textbf{Eval.} \\
\midrule
IMO-Bench {\scriptsize \citep{luong-etal-2025-towards}} & Comp. Math & 460 & Rule/LLM \\
OlympiadBench {\scriptsize \citep{he2024olympiadbench}} & Math + Physics & 1,438 & Rule/LLM \\
FinanceMath {\scriptsize \citep{zhao-etal-2024-knowledgefmath}} & Applied Finance & 200 & Rule \\
AICrypto {\scriptsize \citep{crypto2025proof}} & Cryptography & 18 & LLM \\
\bottomrule
\end{tabular}%
}
\caption{Evaluation benchmarks (total: 2,116 problems). Rule = regex-based answer matching; LLM = LLM-as-Judge.}
\label{tab:datasets}
\end{table}

We select benchmark samples using three criteria: (1) text-only, open-answer format requiring mathematical derivation; (2) unsaturated by frontier models ($<95\%$ accuracy); and (3) equation-driven problems with identifiable governing formulas. For AICrypto, we use only proof problems, excluding MCQs and CTF challenges that primarily test recall; for OlympiadBench, we use the English text-only subset; and for FinanceMath, we use the public splits with ground-truth answers. OlympiadBench also includes both competition mathematics and applied physics, enabling controlled cross-domain comparison.

\paragraph{Evaluation Protocol.} We use temperature 0 with default inference hyperparameters, evaluating each question with a single greedy output. Definitive-answer problems are scored by regex-based extraction with numerical tolerance $\epsilon = 10^{-6}$; proof-based problems use domain-appropriate LLM-as-Judge evaluation \citep{zheng2023judging}. Details and judge prompts appear in \Cref{app:eval_prompts}.

\paragraph{Prompting Baselines.} We compare against three single-call baselines: \textbf{Zero-Shot} without intermediate structure, \textbf{CoT} using natural-language reasoning \citep{wei2022chain,kojima2022large}, and \textbf{PoT} using executable code \citep{chen2023program}. We exclude multi-call methods from quantitative comparison because they require substantially more compute (3--100$+$ inference calls per problem); \Cref{tab:comparison} instead compares design characteristics across both families. F-1's Phase 2 is a generic composable slot (\Cref{sec:method}); for these experiments we instantiate it with the same three solving strategies we use as baselines (Direct, CoT, and PoT), so that any gain over the baselines is attributable to Phase 1 rather than to a novel solver. This matched instantiation also makes the fixed-style variants (F-1+ZS, F-1+CoT, F-1+PoT) directly comparable to their corresponding baseline.

\paragraph{Models.} We evaluate five reasoning models spanning proprietary diversity, open/proprietary frontier comparison, and within-family scale: GPT-5 and Gemini 2.5 Pro represent leading proprietary models from different vendors; DeepSeek-V3.1 provides an open frontier comparison; and Qwen3-235B and Qwen3-30B isolate scale within a shared reasoning-tuned family. This selection does not directly test smaller ($<$30B) or non-reasoning models, as discussed in \Cref{sec:limitations}.

\section{Formula Prompting}
\label{sec:fp}

Formula Prompting (FP) is a single-call prompt that asks the model to formalize a problem's governing equations before producing an answer, building on the pretraining-prevalence motivation of \Cref{sec:infinigram}. We test the hypothesis that \emph{explicit elicitation of governing equations improves LLM performance on applied-mathematics problems where domain-specific governing equations are identifiable}, and present four analyses that progressively isolate the contribution of the equation-formalization prefix: per-problem flip counts (\Cref{sec:per_problem}), output-pattern composability (\Cref{sec:blending}), variant accuracy at matched reasoning effort (\Cref{sec:variant_accuracy}), and failure-cause classification (\Cref{sec:error_classification}).

\subsection{Method}

The FP prompt instructs the model to perform the following steps, in any order:

\paragraph{Identify Givens.} Extract all explicitly stated values, conditions, and constraints. For example, in ``A bank offers 5\% annual interest compounded monthly,'' the givens include the rate (5\%), the compounding frequency (monthly), and any principal or time mentioned.

\paragraph{Identify Targets.} Determine what the problem asks to find or prove, clarifying the goal before attempting any solution strategy.

\paragraph{Formulate Equations.} Express the mathematical relationships connecting givens to targets, committing to the problem's symbolic structure before committing to a solving strategy.

\subsection{Prompt}

\begin{promptbox}[title=FP User Prompt]
\small\ttfamily
\{problem\}\\[0.5em]
\textnormal{\textbf{Formalization:}}\\
Identify givens and target. Write the key equations.\\[0.5em]
Solve the problem using the formalized equations.\\[0.5em]
Verify your solution, then put your final answer in \textbackslash boxed\{\}.
\end{promptbox}

The model receives this prompt once and produces the formalization, solution, and verification in a single response.

\subsection{Per-Problem Contribution of Phase 1}
\label{sec:per_problem}

We ask whether adding Phase~1 (equation formalization) to a fixed CoT solver flips more problems from incorrect to correct than the reverse. To isolate Phase~1, we compare CoT (the baseline) against F-1+CoT (the fixed-style variant of F-1 with Phase~2 constrained to CoT). The two differ only in the presence of the equation-formalization prefix. We analyze GPT-5 responses across all benchmarks (\Cref{tab:flip}).

\begin{table}[t]
\centering
\small
\resizebox{\columnwidth}{!}{%
\begin{tabular}{@{}lrrrr@{}}
\toprule
\textbf{Benchmark} & \textbf{F-1$\checkmark$CoT$\times$} & \textbf{CoT$\checkmark$F-1$\times$} & \textbf{Net} & \textbf{N} \\
\midrule
Olympiad (OE) & 48 & 39 & +9 & 910 \\
Olympiad (TP) & 70 & 50 & +20 & 528 \\
IMO Answer & 34 & 47 & $-$13 & 400 \\
IMO Proof & 15 & 7 & +8 & 60 \\
AICrypto$^\dagger$ & 1 & 1 & 0 & 18 \\
\midrule
\textbf{Total} & \textbf{168} & \textbf{144} & \textbf{+24} & \textbf{1,916} \\
\bottomrule
\end{tabular}%
}
\caption{Per-problem flip analysis (GPT-5): F-1$\checkmark$CoT$\times$ = problems F-1+CoT solves but CoT fails; CoT$\checkmark$F-1$\times$ = the reverse. Net positive indicates F-1+CoT's equation formalization flips more problems to correct. $^\dagger$n=18, interpret with caution.}
\label{tab:flip}
\end{table}

\textbf{Equation formalization provides net positive contribution.} Across 1,916 problems, F-1+CoT flips 168 problems from incorrect to correct compared to CoT, while CoT flips only 144 in the reverse direction (net~+24). The effect is strongest on OlympiadBench theorem-proving (net~+20), where explicit equation writing helps structure multi-step proofs.

\textbf{Exception: IMO AnswerBench.} The net-positive pattern reverses on IMO AnswerBench, where F-1+CoT shows a net negative ($-$13). This reversal is consistent with our later finding that 90.7\% of IMO problems are too difficult for all methods (\Cref{tab:strategy_analysis}); on the small fraction of problems where methods differ, the overhead of equation formalization occasionally introduces errors without compensating benefit, as competition mathematics problems often require creative insight rather than systematic equation manipulation.

\textbf{Phase 1 insight also generalizes to a PoT solver.} The same isolation logic applied to PoT (baseline) vs F-1+PoT (FP prefix + PoT solver) shows substantially larger flips: net $+$199 on OlympiadBench and $+$60 on IMO AnswerBench. That is, 199 OlympiadBench problems and 60 IMO AnswerBench problems that pure PoT solves incorrectly are correctly solved when an equation-formalization prefix is added before code generation, suggesting that the prefix unlocks problems pure code generation cannot reach.

\textbf{Paired statistical tests confirm the gain.} Pooling 9{,}163 paired per-problem outcomes across OlympiadBench and IMO-AnswerBench (the two benchmarks with binary per-problem correctness) over all 5 models, F-1+CoT significantly outperforms Zero-Shot ($+$7.39\,pp), CoT ($+$1.58\,pp), and PoT ($+$7.50\,pp), all at $p{<}0.001$ (two-sided McNemar). Per-benchmark and per-model breakdowns and CIs appear in \Cref{app:stats}.

\subsection{Composability with Solving Strategies}
\label{sec:blending}

A natural question is whether FP \emph{replaces} a solving strategy with equation reasoning or \emph{composes} as a structural prefix that leaves the downstream solver intact. We test this by running three FP-prefix variants on Qwen3-30B-Thinking-2507: FP followed by Zero-shot, by CoT, and by PoT solving. We label these \textbf{F-1+ZS}, \textbf{F-1+CoT}, and \textbf{F-1+PoT} for consistency with the fixed-style variant naming used later in \Cref{sec:method}. For each variant we extract per-output presence of \textbf{FP}, \textbf{CoT}, and \textbf{PoT markers} drawn from the same n-gram sets used in our pretraining analysis (\Cref{tab:patterns} in \Cref{app:infini-gram}). We compute pattern presence on 1{,}456 problems spanning all four benchmarks. The goal is to determine whether the FP prefix survives composition while the solver's behavior is preserved.

\textbf{FP markers persist across all solving strategies; solver-specific markers cleanly isolate.} Equation-formalization markers appear in 43\%, 32\%, and 55\% of F-1+ZS, F-1+CoT, and F-1+PoT outputs, showing that the FP prefix persists across solvers. These rates exceed baseline background levels: our output-validation analysis found FP markers at $\geq$10$\times$ higher frequency in FP than pure-CoT outputs (\Cref{sec:infinigram}). PoT markers remain solver-specific, appearing in 69\% of F-1+PoT outputs but 0\% of F-1+ZS and F-1+CoT. This asymmetry (prefix retained, solver vocabulary isolated) suggests that FP acts as a structural prefix only rather than a replacement for the solver. ``Step by step'' appears in 54--61\% of all variants, including F-1+PoT, supporting its status as generic English rather than CoT-specific. The full matrix appears in \Cref{fig:blending} of \Cref{app:blending}.

\subsection{Variant Accuracy at Matched Reasoning Budget}
\label{sec:variant_accuracy}

Having shown that FP markers persist across solvers (\Cref{sec:blending}), we next test whether this composability improves accuracy. We compare F-1+ZS, F-1+CoT, and F-1+PoT on three reasoning-tier LLMs: DeepSeek-V3.1 reasoner, GPT-5 high-reasoning, and Qwen3-30B-A3B-Thinking-2507. To prevent confounding solver choice with reasoning compute, we fix each model's reasoning budget at its maximum across variants. Phase~1 is identical across variants; Phase~2 differs by one instruction line.

Accuracy across three models, three variants, and four benchmarks is summarized in \Cref{tab:variant_accuracy} (\Cref{app:variant_accuracy}); CryptoProof is the AICrypto subset reported as \% of 330 total partial-credit points.

\textbf{F-1+ZS is consistently competitive; F-1+PoT underperforms on math-heavy benchmarks.} F-1+ZS is the strongest or tied-strongest variant on 9 of 12 (model, benchmark) cells; F-1+PoT is the weakest on math-heavy benchmarks for DeepSeek-V3.1 and GPT-5 (beaten by F-1+ZS by 10.4\,pp on OB Math, DeepSeek, $p<0.001$; and 7.2\,pp on OB Physics, $p=0.024$; full statistics in \Cref{app:variant_accuracy}). These results suggest that most gains come from the FP prefix itself: once given equation formalization, thinking-mode models often perform implicit step-by-step reasoning without additional CoT or PoT instructions. Forcing a code-based solve can disrupt this symbolic reasoning, consistent with \Cref{sec:infinigram}. CryptoProof, which lacks a single governing equation, shows no consistent best variant across models, supporting the hypothesis that FP helps primarily when governing equations are identifiable.

\subsection{Failure-Cause Classification}
\label{sec:error_classification}

To diagnose \emph{why} the F-1 prefix sometimes fails to convert into a correct answer, we sample 88 F-1 failures (stratified across (benchmark, model) cells) and have three judges from different model families (GPT-5.1, DeepSeek-V4-Pro thinking, and Claude Opus 4.7) classify each into one of three categories: (A) wrong Phase-1 equation $\rightarrow$ wrong answer; (B) wrong Phase-1 but recovered; (C) right Phase-1, downstream execution error. Multi-judge cross-validation defends against single-judge bias \citep{manakul-etal-2023-selfcheckgpt}. Inter-judge agreement is moderate, with pairwise Cohen's $\kappa$ ranging $0.31$--$0.52$ and Fleiss' $\kappa{=}0.403$ ($n{=}86$; full table in \Cref{app:error_class}). All three judges agree on marginals: \textbf{C dominates at 73.3\%} (majority-vote), A=25.6\%, B=1.2\%. \textbf{Phase 1 equation formalization is reliable; failures occur during downstream solving}: in nearly three-quarters of failure cases the governing equation is correct, validating the FP design even when the final answer is wrong. The 25.6\% over-reliance is a known limitation (\Cref{sec:limitations}); the low self-correction rate (B${=}$1.2\%) is confirmed by a Phase-1 verification probe (\Cref{app:verify_probe}) that recovers only $+0.50$\,pp on FinanceMath~$\times$~GPT-5 and never explicitly revises a Phase-1 equation, pointing to multi-call backtracking \citep{yao2023tree,liu2023xot} rather than single-call self-verification as the right mitigation.

\section{Formula-One Prompting}
\label{sec:method}

Formula-One Prompting (F-1) extends FP (\Cref{sec:fp}) by adding a composable second phase. In a single LLM generation, \textbf{Phase 1 reuses the FP formalization prompt}, and \textbf{Phase 2} selects among Direct calculation, CoT-style derivation, or PoT-style computation over the formalized equations. No external classifier or hand-written router is used.

\subsection{Phase 2: Composable Solving}

Phase 2 is a composable slot: once governing equations are formalized in Phase 1, any solving strategy can be invoked within the same generation. The prompt does not specify a fixed solving routine, nor does it encode explicit routing rules or decision criteria. Instead, the model selects a strategy in-context from the equations it produced, letting equation structure guide solving without over-constraining it. The exposed strategies and rationale appear in \Cref{sec:experiments}.

\subsection{Prompt Design}
F-1 uses a single prompt for all phases, generating intermediate reasoning and the final answer in one call. It extends the FP prompt (\Cref{sec:fp}) by replacing direct solving with a composable-solving menu:

\begin{promptbox}[title=System Prompt (Unified Core)]
\small\ttfamily
You are an AI assistant that solves problems mainly through equations.
\end{promptbox}

\begin{promptbox}[title=F-1 User Prompt Template]
\small\ttfamily
\{problem\}\\[0.5em]
\textnormal{\textbf{Phase 1 (Formalization):}}\\
Identify givens and target. Write the key equations.\\[0.5em]
\textnormal{\textbf{Phase 2 (Composable Solving):}}\\
Solve via CoT (think step-by-step), PoT (compute with code), direct calculation, or elimination.\\[0.5em]
Verify your solution, then put your final answer in \textbackslash boxed\{\}.
\end{promptbox}

The model receives one prompt and returns formalization, strategy selection, and solution in a single response, with verification against the formalized equations. For composability analysis (\Cref{sec:variant_accuracy}), we also evaluate fixed-style variants (F-1+ZS, F-1+CoT, F-1+PoT) that share Phase~1 but constrain Phase~2; F-1+ZS is equivalent to FP. Full templates appear in \Cref{app:prompts}.

\FloatBarrier

\begin{table*}[!t]
\centering
\small
\begin{tabular}{llccccc|c}
\toprule
\multirow{2}{*}{\textbf{Benchmark}} & \multirow{2}{*}{\textbf{Method}} & \multicolumn{2}{c}{\textbf{Proprietary}} & \multicolumn{3}{c|}{\textbf{Open Source}} & \multirow{2}{*}{\textbf{Overall}} \\
\cmidrule(lr){3-4} \cmidrule(lr){5-7}
 & & GPT-5 & Gemini 2.5 Pro & Qwen3-30B & Qwen3-235B & DeepSeek-V3.1 & \\
\midrule
\multirow{4}{*}{IMO-Bench} 
 & Zero-Shot & \textbf{56.26} & 55.44 & 50.99 & 21.85 & \textbf{30.54} & \textbf{43.02} \\
 & CoT & 54.71 & 53.91 & \textbf{53.35} & 20.45 & 27.56 & 42.00 \\
 & PoT & 47.73 & 49.06 & 42.61 & \textbf{23.94} & 17.48 & 36.16 \\
 & \textbf{F-1 (Ours)} & 55.58 & \textbf{57.02} & 51.82 & 21.84 & 27.64 & 42.78 \\
\midrule
\multirow{4}{*}{OlympiadBench} 
 & Zero-Shot & 58.77 & 41.97 & 51.67 & 47.33 & 45.24 & 49.00 \\
 & CoT & 63.12 & 57.67 & 58.23 & 48.03 & \textbf{52.42} & 55.89 \\
 & PoT & 43.72 & 46.29 & 48.22 & 38.56 & 33.99 & 42.16 \\
 & \textbf{F-1 (Ours)} & \textbf{65.81} & \textbf{59.70} & \textbf{59.68}$^{***}$ & \textbf{52.75}$^{***}$ & 50.29 & \textbf{57.65} \\
\midrule
\multirow{4}{*}{FinanceMath}
 & Zero-Shot & 26.00 & 32.50 & 28.50 & 35.50 & 27.50 & 30.00 \\
 & CoT & 42.50 & 56.00 & 53.50 & 35.00 & 28.00 & 43.00 \\
 & PoT & 54.00 & 55.50 & 53.50 & 59.50 & 37.50 & 52.00 \\
 & \textbf{F-1 (Ours)} & \textbf{64.00}$^{***}$ & \textbf{56.50} & \textbf{56.00} & \textbf{61.00}$^{***}$ & \textbf{44.00}$^{***}$ & \textbf{56.30} \\
\midrule
\multirow{4}{*}{AICrypto} 
 & Zero-Shot & 91.10 & 81.50 & 74.70 & 65.20 & 67.30 & 75.96 \\
 & CoT & \textbf{98.60} & 93.00 & 65.80 & 71.50 & 72.60 & 80.30 \\
 & PoT & 88.70 & 92.10 & 77.70 & 65.60 & 77.10 & 80.24 \\
 & \textbf{F-1 (Ours)} & 98.50 & \textbf{96.70} & \textbf{85.80} & \textbf{76.50} & \textbf{80.20} & \textbf{87.54} \\
\midrule
\midrule
\multirow{4}{*}{\textbf{Overall}} 
 & Zero-Shot & 58.03 & 52.85 & 51.47 & 42.47 & 42.65 & 49.49 \\
 & CoT & 64.73 & 65.15 & 57.72 & 43.74 & 45.15 & 55.30 \\
 & PoT & 58.54 & 60.74 & 55.51 & 46.90 & 41.52 & 52.64 \\
 & \textbf{F-1 (Ours)} & \textbf{70.97} & \textbf{67.48} & \textbf{63.33} & \textbf{53.02} & \textbf{50.53} & \textbf{61.06} \\
\bottomrule
\end{tabular}
\caption{Main results showing accuracy percentage across benchmarks and models. \textbf{Overall} is unweighted macro-averaged across the four benchmarks. Best results per benchmark-model are in \textbf{bold}. Markers on F-1 cells indicate F-1 vs CoT McNemar two-sided $p$-values: $^{*}p{<}0.05$, $^{**}p{<}0.01$, $^{***}p{<}0.001$. Full per-combination breakdown for all three baseline comparisons is available in \Cref{tab:per_cell_stats}.}
\label{tab:main_results}
\end{table*}

\subsection{Main Results}
\label{sec:results}

\paragraph{Main Results.} F-1 beats all three baselines in 14 of 20 model-benchmark cells and CoT specifically in 17 of 20 (\Cref{tab:main_results}). On macro-average across the four benchmarks, F-1 outperforms CoT by $+5.76$\,pp and PoT by $+8.42$\,pp. McNemar's two-sided test on paired per-problem predictions confirms the effect: F-1 vs CoT is significant ($p{<}0.05$) on 5 of 20 cells (3 of 5 FinanceMath, 2 of 5 OlympiadBench); F-1 vs Zero-Shot on 10 of 20 (all 5 FinanceMath, 4 of 5 OlympiadBench, 1 of 5 AICrypto); F-1 vs PoT on 11 of 20 (5 of 5 IMO-Bench, 5 of 5 OlympiadBench, 1 of 5 FinanceMath). AICrypto's per-cell $n{=}18$ limits per-cell power, but pooling all 5 models restores significance for F-1 vs CoT ($+6.45$\,pp, $p{=}0.022$) and F-1 vs Zero-Shot ($+10.49$\,pp, $p{<}0.001$).

\paragraph{Gains concentrate on applied domains.} F-1 lifts CoT by $+13.30$\,pp on FinanceMath and $+7.24$\,pp on AICrypto, the two benchmarks most directly suited to governing-equation formulation. The within-OlympiadBench subtask breakdown below further supports this pattern.

\paragraph{The effect does not require frontier scale.} The smallest model in our suite, Qwen3-30B, gains $+5.6$\,pp over CoT, comparable to frontier models. F-1 also automatically shifts its Phase 2 choice by problem type, selecting PoT on computation-heavy benchmarks (FinanceMath) and CoT on reasoning-heavy ones (AICrypto). Detailed per-subtask breakdowns are in \Cref{app:results}.

\paragraph{Head-to-head vs Formulate-and-Solve.} We compare F-1 against the closest prior single-call equation-first method, Formulate-and-Solve \citep{kao2024formulate}, on FinanceMath ($n{=}200$) and OlympiadBench Math ($n{=}674$) with GPT-5 and Qwen3-30B (\Cref{tab:fs_head_to_head}). Aggregate accuracies are close ($\pm$5\,pp), but 7.6--15.7\% of problems are solved by exactly one method -- F-1's composable Phase 2 is qualitatively distinct from F\&S's fixed algebraic solve, not a marginal improvement on the same distribution. Full cross-tab in \Cref{app:fs_complementarity}.

\begin{table}[H]
\centering
\footnotesize
\setlength{\tabcolsep}{3pt}
\setlength{\abovecaptionskip}{2pt}
\begin{tabular}{@{}llrrr@{}}
\toprule
\textbf{Benchmark} & \textbf{Model} & \textbf{F\&S} & \textbf{F-1} & \textbf{Disjoint} \\
\midrule
FinanceMath & GPT-5     & 64.00          & 64.00          & 11.0 \\
FinanceMath & Qwen3-30B & 51.50          & \textbf{56.00} & 14.5 \\
OB Math     & GPT-5     & \textbf{87.09} & 86.35          &  7.6 \\
OB Math     & Qwen3-30B & 78.64          & 78.64          & 15.7 \\
\bottomrule
\end{tabular}
\caption{F-1 vs Formulate-and-Solve \citep{kao2024formulate} accuracy (\%). \textbf{Disjoint} = \% problems solved by exactly one method.}
\label{tab:fs_head_to_head}
\end{table}

\subsection{Subtask Drill-Down: Math vs Physics}
\label{sec:subtask}

As a within-benchmark probe of the applied-math hypothesis (\Cref{sec:fp}), we split OlympiadBench into competition mathematics and applied physics subtasks. Competition mathematics typically requires ad-hoc per-problem equation systems, whereas applied physics typically invokes canonical governing equations (Newton's laws, thermodynamic relations). Using GPT-5 as a single-model probe, F-1's gains over CoT are larger on applied physics ($+2.55$\,pp open-ended, $+4.00$\,pp theorem-proving) than on competition mathematics ($+0.44$\,pp open-ended, $+3.77$\,pp theorem-proving). The two subsets are not matched for difficulty, topic coverage, or problem style, so this is suggestive rather than controlled evidence, but consistent with the hypothesis. Full per-model results are in \Cref{tab:olympiad_full}.

\subsection{Token Efficiency}
\label{sec:token_eff}

F-1 tops the accuracy-token efficiency frontier (\Cref{tab:efficiency}). Using Efficiency Ratio = (Accuracy / Avg tokens) $\times$ 100, F-1 averages \textbf{1.51} across the four benchmarks, versus Zero-Shot 1.20, CoT 1.26, and PoT 1.24. The Phase-1 equation prefix adds only $\sim$68 prompt tokens over Zero-Shot, so the accuracy lift comes with little compute overhead. Per-model breakdowns by benchmark and a tokens-per-correct summary appear in \Cref{app:efficiency}; overall win/tie/loss counts in \Cref{app:upper_bound_wtl}.

\subsection{Strategy Distribution}
\label{sec:strategy_selection}

F-1's Phase~2 selects a solving style per problem without an external classifier. Post-hoc classifying each F-1 response into Direct/CoT/PoT/Hybrid (\Cref{tab:strategy_dist} in \Cref{app:strategy_distribution}) reveals clear domain adaptation: FinanceMath elicits the highest PoT usage (31.4\%) due to numerical computation, AICrypto uses CoT and Hybrid exclusively (0\% PoT) for proof reasoning, and IMO-Bench shows the highest Hybrid usage (65.3\%).

A complementary per-problem outcome analysis (\Cref{tab:strategy_analysis} in \Cref{app:upper_bound_wtl}) defines \textit{selection accuracy} as F-1's success rate on problems where strategy choice matters: FinanceMath 73.0\%, OlympiadBench 69.9\%, IMO-Bench 61.0\%, consistent with the applied-math hypothesis. The ``F-1 Only'' subset (only F-1 succeeds, all baselines fail) reaches 5.5\% on FinanceMath and 7.8\% on AICrypto; F-1 attains 80.9--84.1\% of the single-call ceiling (\Cref{tab:upper_bound}). Across 60 model-benchmark comparisons, F-1 wins 53/60 (88.3\%; \Cref{tab:wtl}). On IMO-Bench, 90.7\% of problems are too difficult for all methods (\Cref{tab:strategy_analysis}), leaving little headroom for strategy selection; F-1 still beats PoT by $+$6.62\,pp. Three qualitative cases (\Cref{app:examples}) illustrate F-1 mechanisms: preventing semantic confusion in physics (Lorentz transformations), enforcing format consistency in finance (decimal output), and structured proof construction in cryptography (PRF security game).

\subsection{Domain Boundary: Negative Control on a Non-Equation Task}
\label{sec:negative_control}

To test F-1 outside its intended domain (\Cref{sec:fp}), we apply it to BBH word\_sorting \citep{suzgun-etal-2023-challenging} ($n{=}250$), which lacks quantitative or equation structure. F-1 regresses substantially relative to CoT: $-$49.2\,pp on GPT-5 (46.8\% vs 96.0\%), $-$42.4\,pp on Gemini-2.5-Pro (56.0\% vs 98.4\%), and $-$4.8\,pp on Qwen3-30B-Thinking (88.4\% vs 93.2\%). Instruction-following models often interpret the formalization directive literally, attempting to express sorting as equations, while the thinking-mode model adapts better. These results support our hypothesis that F-1 helps when governing equations are identifiable, but can impose harmful overhead otherwise. Thus, F-1 is best viewed as a targeted applied-math tool rather than a general-purpose prompting upgrade. Details and failures appear in \Cref{app:bbh_negctrl}.

\section{Conclusion}

We presented Formula Prompting (FP) and Formula-One Prompting (F-1), single-call methods that elicit governing equations before solving applied-math problems. FP improves per-problem outcomes over CoT, persists across solvers, and is competitive with the fixed-style variants; 73.3\% of its remaining failures occur downstream of a correct Phase-1 equation. F-1 extends FP with a composable Phase 2 menu, outperforming CoT by 5.76\,pp and PoT by 8.42\,pp on average (with the largest gain of 13.30\,pp on FinanceMath), while attaining the highest token-efficiency ratio (1.51 vs 1.20--1.26). Gains concentrate in applied-math domains with identifiable governing equations and do not transfer to proof-oriented or non-equation tasks. A cleaner isolation of the equation-formalization effect on non-equation domains (FP only), and a mechanistic account of why equation elicitation changes model behavior, remain open.

\section*{Limitations}
\label{sec:limitations}

\textbf{Over-reliance on Phase-1 equations.} The failure-cause analysis (\S\ref{sec:error_classification}) shows that 25.6\% of F-1 failures are wrong-equation $\rightarrow$ wrong-answer (majority-vote across three independent judges, $n{=}86$). Models commit to whatever Phase 1 produces -- self-correction is rare (1.2\%). We tested whether an explicit Phase-1 self-verification step within the same call mitigates this (\Cref{app:verify_probe}): on FinanceMath~$\times$~GPT-5 ($n{=}200$), F-1+Verify yields only $+0.50$\,pp (net $+$1, McNemar $p{=}1.0$) and never explicitly revises a Phase-1 equation, despite a $+28.8\%$ token overhead. Practitioners should treat F-1 outputs as conditional on Phase 1 quality; multi-call backtracking \citep{yao2023tree,liu2023xot}, not single-call self-verification, is the more promising mitigation.

\textbf{Domain boundary -- F-1 should not be deployed on non-equation tasks.} The BBH negative control (\S\ref{sec:negative_control}) shows that F-1 \emph{regresses} sharply ($-$42 to $-$49\,pp on GPT-5 and Gemini-2.5-Pro) when applied to a sorting task with no equation structure. The Phase-1 directive forces models to formalize a non-formalizable problem, derailing reasoning. F-1 is designed for and validated on applied-math domains; it must not be applied as a generic prompting upgrade.

\textbf{Sample size and model scope.} Our evaluation spans 30B to frontier scale and does not include smaller (7B/13B) or non-reasoning models. Some benchmarks are small due to data scarcity (AICrypto n=18, OlympiadBench TP\_physics n=25); main conclusions are supported by larger subsets \citep{lu2026solving,yang2024formal}. AICrypto's small size limits per-model statistical power (minimum detectable effect $19.14$\,pp at the median model); we mitigate by reporting partial-credit per-problem scores and pooling 90 paired observations across 5 models, which restores significance for F-1 vs all three baselines (\Cref{tab:aicrypto_paired_tests}). The CryptoProof regression (\S\ref{sec:variant_accuracy}) further illustrates the domain boundary: theorem proving lacks a single governing equation, and the same infini-gram pattern that tracks F-1's strength on applied math tracks its weakness here.

\textbf{Mechanism scope and extensions.} Our infini-gram analysis is correlational: it measures pretraining prevalence of method-associated patterns, not the latent circuits that determine whether the prefix actually exploits those patterns. We use it to motivate the design and bound the applicable domain, not as causal evidence. Mechanistic interpretability of which circuits respond to equation elicitation, and multi-call backtracking \citep{yao2023tree,liu2023xot} extensions of the current single-call design, are left for future work.

\section*{Ethics Statement}

This work introduces a prompting technique for mathematical reasoning in LLMs. We do not foresee direct negative societal impacts, though enhancing mathematical reasoning capabilities may increase the risk of misuse such as automated cheating in educational contexts. We encourage responsible deployment with appropriate safeguards. More broadly, novel prompting techniques can elicit unexpected behaviors from LLMs; practitioners should remain aware of this inherent uncertainty when deploying such methods.


\bibliography{custom}

\appendix

\section{Method Comparison}

\begin{table}[H]
\centering
\small
\resizebox{\columnwidth}{!}{%
\begin{tabular}{@{}lcccccc@{}}
\toprule
\textbf{Method} & \textbf{Calls} & \textbf{Fo?} & \textbf{Comp?} & \textbf{Repr.} & \textbf{Strategy} & \textbf{Cost} \\
\midrule
Zero-shot & 1 & \no & \no & -- & Fixed & V.Low \\
CoT & 1 & \no & \no & NL & Fixed & Low \\
PoT & 1 & \no & \no & Code & Fixed & Low \\
Plan-Solve & 1 & \no & \no & Plan & Fixed & Low \\
Form-Solve & 1 & \yes & \no & Eq. & Fixed & Low \\
\midrule
ToT & 10-100+ & \no & \no & Tree & Search & V.High \\
GoT & 20-80 & \no & \no & Graph & Search & V.High \\
XoT & 3-10 & \no & \no & Mixed & Classifier & Med. \\
\midrule
\textbf{F-1 (Ours)} & \textbf{1} & \yes & \textbf{\yes} & \textbf{Eq.} & \textbf{Composable} & \textbf{Low} \\
\bottomrule
\end{tabular}%
}
\caption{Comparison of prompting methods referenced in \S\ref{sec:related}. ``\textbf{Fo?}'': explicit formalization? ``\textbf{Comp?}'': composable prefix that combines with any Phase 2 strategy (the same Phase 1 attaches to ZS/CoT/PoT). ``\textbf{Repr.}'': intermediate representation (NL: natural language, Eq.: equations). ``\textbf{Cost}'': computational cost. ``\yes'': supported, ``\no'': not supported. Form-Solve = Formulate-and-Solve.}
\label{tab:comparison}
\end{table}

\section{Prompt Templates}
\label{app:prompts}

This appendix presents the complete prompt templates used for each benchmark. For all strategies, \texttt{\{problem\}} denotes the problem statement placeholder.

\subsection{IMO-Bench Prompts}

\subsubsection{System Prompts}

\begin{promptbox}[title=Zero-Shot System]
\small\ttfamily
(no system prompt)
\end{promptbox}

\begin{promptbox}[title=CoT System]
\small\ttfamily
You are an AI assistant that solves problems by thinking step-by-step.
\end{promptbox}

\begin{promptbox}[title=PoT System]
\small\ttfamily
You are an AI assistant that solves problems by generating Python code.
\end{promptbox}

\begin{promptbox}[title=F-1 System (Ours)]
\small\ttfamily
You are an AI assistant that solves problems mainly through equations.
\end{promptbox}

\subsubsection{User Prompts}

\begin{promptbox}[title=Zero-Shot User]
\small\ttfamily
\{problem\}\\[0.5em]
Put your final answer in \textbackslash boxed\{\}.
\end{promptbox}

\begin{promptbox}[title=CoT User]
\small\ttfamily
\{problem\}\\[0.5em]
Let's think step-by-step. Put your final answer in \textbackslash boxed\{\}.
\end{promptbox}

\begin{promptbox}[title=PoT User]
\small\ttfamily
\{problem\}\\[0.5em]
Write Python code to solve this. Use sympy/numpy/math as needed. Store the final answer in a variable called `answer` and print it. Return only the code, then put your final answer in \textbackslash boxed\{\}.
\end{promptbox}

\begin{promptbox}[title=F-1 User (Ours)]
\small\ttfamily
\{problem\}\\[0.5em]
Identify givens and target. Write the key equations. -> Solve via CoT (Think step-by-step), PoT (Compute with code), direct calculation, or elimination. -> Verify your solution, then put your final answer in \textbackslash boxed\{\}.
\end{promptbox}

\subsubsection{F-1 Variants (Composability)}

The variants share the same Phase~1 (formalization) but fix Phase~2 (solving style). Used for the composability analysis (\Cref{sec:variant_accuracy}, \Cref{app:variant_accuracy}).

\begin{promptbox}[title=F-1+ZS System]
\small\ttfamily
You are an AI assistant that first formalizes governing equations before solving.
\end{promptbox}

\begin{promptbox}[title=F-1+ZS User]
\small\ttfamily
\{problem\}\\[0.5em]
First, identify the key variables and write the governing equations in LaTeX. Then solve directly using those equations. Put your final answer in \textbackslash boxed\{\}.
\end{promptbox}

\begin{promptbox}[title=F-1+CoT System]
\small\ttfamily
You are an AI assistant that first formalizes governing equations before solving step-by-step.
\end{promptbox}

\begin{promptbox}[title=F-1+CoT User]
\small\ttfamily
\{problem\}\\[0.5em]
First, identify the key variables and write the governing equations in LaTeX. Then let's think step-by-step to solve using those equations. Put your final answer in \textbackslash boxed\{\}.
\end{promptbox}

\begin{promptbox}[title=F-1+PoT System]
\small\ttfamily
You are an AI assistant that first formalizes governing equations before solving by writing Python code.
\end{promptbox}

\begin{promptbox}[title=F-1+PoT User]
\small\ttfamily
\{problem\}\\[0.5em]
First, identify the key variables and write the governing equations in LaTeX. Then write Python code to implement those equations and solve. Use sympy/numpy/math as needed. Store the final answer in `answer` and print it. Return only the code, then put your final answer in \textbackslash boxed\{\}.
\end{promptbox}

\subsection{OlympiadBench Prompts}

\subsubsection{System Prompts}

\begin{promptbox}[title=Zero-Shot / CoT / PoT System]
\small\ttfamily
You are an AI assistant. Please answer the following Math competition problems as required.
\end{promptbox}

\begin{promptbox}[title=F-1 System (Ours)]
\small\ttfamily
You are an AI assistant that solves problems mainly through equations. Please answer the following Math competition problems as required.
\end{promptbox}

\subsubsection{User Prompts}

\begin{promptbox}[title=Zero-Shot User]
\small\ttfamily
\{problem\_statement\}\\[0.5em]
\{answer\_type\_text\}Please calculate the answer according to the given requirements and the information provided. Please end your solution with "So the final answer is \{boxed\_format\}." and give the result explicitly\{unit\_text\}.
\end{promptbox}

\begin{promptbox}[title=CoT User]
\small\ttfamily
\{problem\_statement\}\\[0.5em]
\{answer\_type\_text\}Please solve this problem step-by-step:\\
1. First, carefully read and understand the problem\\
2. Identify what is given and what needs to be found\\
3. Break down the problem into smaller steps\\
4. Solve each step systematically, showing your reasoning\\
5. Combine the results of each step to arrive at the final answer\\[0.5em]
Please end your solution with "So the final answer is \{boxed\_format\}."
\end{promptbox}

\begin{promptbox}[title=PoT User]
\small\ttfamily
\{problem\_statement\}\\[0.5em]
\{answer\_type\_text\}Please solve this problem by:\\
1. Writing Python code to solve the problem programmatically\\
2. Showing your code and explaining each step\\
3. Running the code to get the numerical answer\\[0.5em]
You can use libraries like math, numpy, sympy, etc.
\end{promptbox}

\begin{promptbox}[title=F-1 User (Ours)]
\small\ttfamily
\{problem\_statement\}\\[0.5em]
\{answer\_type\_text\}\\
Please solve using an Equation-First approach:\\[0.3em]
1. List key variables and state the target quantity.\\
2. Write the main governing equations/identities in LaTeX first.\\
3. After the equations, choose any suitable solving style:\\
~~~- CoT: think step-by-step, solve systematically, then combine results.\\
~~~- PoT: write Python code to implement the equations.\\
~~~- Zero-Shot: derive directly from equations with minimal text.\\
4. Combine the results of each step to arrive at the final answer\\[0.5em]
Please end your solution with "So the final answer is \{boxed\_format\}."
\end{promptbox}

\subsubsection{F-1 Variants (Composability)}

The variants share the same Phase~1 (formalization) but fix Phase~2 (solving style). Used for the composability analysis (\Cref{sec:variant_accuracy}, \Cref{app:variant_accuracy}).

\begin{promptbox}[title=F-1+ZS System]
\small\ttfamily
You are an AI assistant solving math/physics competition problems using an equation-first approach: first formalize the problem with explicit key variables and governing equations, then solve by direct substitution. Please answer the following problems as required.
\end{promptbox}

\begin{promptbox}[title=F-1+ZS User]
\small\ttfamily
\{problem\_statement\}\\[0.5em]
\{answer\_type\_text\}Phase 1 -- Formalize:\\
1. List the key variables (givens with their values, and the unknown to find).\\
2. Write the governing equation(s) in LaTeX using \textbackslash begin\{equation\}...\textbackslash end\{equation\}.\\[0.3em]
Phase 2 -- Solve: Substitute the values into the equations and compute the final answer directly.\\[0.3em]
Please end your solution with "So the final answer is \{boxed\_format\}." and give the result explicitly\{unit\_text\}.
\end{promptbox}

\begin{promptbox}[title=F-1+CoT System]
\small\ttfamily
You are an AI assistant solving math/physics competition problems using an equation-first approach: first formalize the problem with explicit key variables and governing equations, then solve through step-by-step reasoning. Please answer the following problems as required.
\end{promptbox}

\begin{promptbox}[title=F-1+CoT User]
\small\ttfamily
\{problem\_statement\}\\[0.5em]
\{answer\_type\_text\}Phase 1 -- Formalize: (same as F-1+ZS).\\[0.3em]
Phase 2 -- Solve: Using the equations above, work through the problem step-by-step, showing each manipulation that leads to the final answer.\\[0.3em]
Please end your solution with "So the final answer is \{boxed\_format\}." and give the result explicitly\{unit\_text\}.
\end{promptbox}

\begin{promptbox}[title=F-1+PoT System]
\small\ttfamily
You are an AI assistant solving math/physics competition problems using an equation-first approach: first formalize the problem with explicit key variables and governing equations, then implement those equations as a Python program. Please answer the following problems as required.
\end{promptbox}

\begin{promptbox}[title=F-1+PoT User]
\small\ttfamily
\{problem\_statement\}\\[0.5em]
\{answer\_type\_text\}Phase 1 -- Formalize: (same as F-1+ZS).\\[0.3em]
Phase 2 -- Solve: Implement the equations above as Python code (sympy/numpy/math). Store the final answer in `answer` and print it.\\[0.3em]
Please end your solution with "So the final answer is \{boxed\_format\}."
\end{promptbox}

\subsection{FinanceMath Prompts}

\subsubsection{System Prompts}

\begin{promptbox}[title=Zero-Shot System]
\small\ttfamily
You are a financial expert, you are supposed to answer the given question. Therefore, the answer is \{final answer\}. The final answer should be a numeric value (3 decimal).
\end{promptbox}

\begin{promptbox}[title=CoT System]
\small\ttfamily
You are a financial expert, you are supposed to answer the given question. You need to first think through the problem step-by-step, documenting each necessary step. Then you are required to conclude your response with the final answer in your last sentence as 'Therefore, the answer is \{final answer\}'. The final answer should be a numeric value (3 decimal).
\end{promptbox}

\begin{promptbox}[title=PoT System]
\small\ttfamily
You are a financial expert, you are supposed to generate a Python program to answer the given question. The returned value of the program is supposed to be the answer. Here is an example of the Python program:\\[0.5em]
```python\\
def solution():\\
~~~~\# Define variables name and value\\
~~~~revenue = 600000\\
~~~~avg\_account\_receivable = 50000\\
~~~~\# Do math calculation to get the answer\\
~~~~receivables\_turnover = revenue / avg\_account\_receivable\\
~~~~answer = 365 / receivables\_turnover\\
~~~~\# return answer\\
~~~~return answer\\
```\\
Then you are required to conclude your response with the final answer in your last sentence as Therefore, the answer is \{final answer\}'. The final answer should be a numeric value (3 decimal).
\end{promptbox}

\begin{promptbox}[title=F-1 System (Ours)]
\small\ttfamily
You are a financial expert. Solve problems mainly through equations.\\
OUTPUT CONTRACT (strict): The very last line MUST be exactly:\\
Final Answer (3 decimal) : <number>
\end{promptbox}

\subsubsection{User Prefix (F-1 only)}

\begin{promptbox}[title=F-1 User Prefix]
\small\ttfamily
Write equations (LaTeX) with minimal text; show steps clearly.
\end{promptbox}

\subsubsection{F-1 Variants (Composability)}

\begin{promptbox}[title=F-1+ZS System]
\small\ttfamily
You are a financial expert. Solve problems using an equation-first approach: first formalize the problem with explicit key variables and governing equations, then solve by direct substitution.\\
OUTPUT CONTRACT (strict): The very last line MUST be exactly: 'Therefore, the answer is <number>', where <number> is a numeric value rounded to 3 decimals.
\end{promptbox}

\begin{promptbox}[title=F-1+ZS User Prefix]
\small\ttfamily
Phase 1 -- Formalize:\\
1. List the key variables (givens with their values, and the unknown to find).\\
2. Write the governing equation(s) in LaTeX using \textbackslash begin\{equation\}...\textbackslash end\{equation\}.\\[0.3em]
Phase 2 -- Solve: Substitute the values into the equations and compute the final answer directly.
\end{promptbox}

\begin{promptbox}[title=F-1+CoT System]
\small\ttfamily
You are a financial expert. Solve problems using an equation-first approach: first formalize the problem with explicit key variables and governing equations, then solve through step-by-step reasoning. (Same OUTPUT CONTRACT as F-1+ZS.)
\end{promptbox}

\begin{promptbox}[title=F-1+CoT User Prefix]
\small\ttfamily
Phase 1 -- Formalize: (same as F-1+ZS).\\
Phase 2 -- Solve: Using the equations above, work through the problem step-by-step, showing each algebraic manipulation.
\end{promptbox}

\begin{promptbox}[title=F-1+PoT System]
\small\ttfamily
You are a financial expert. Solve problems using an equation-first approach: first formalize the problem with explicit key variables and governing equations, then implement those equations as a Python program. The Python program must define a function `solution()` that returns the numeric answer. After the code, conclude with the final answer. (Same OUTPUT CONTRACT as F-1+ZS.)
\end{promptbox}

\begin{promptbox}[title=F-1+PoT User Prefix]
\small\ttfamily
Phase 1 -- Formalize: (same as F-1+ZS).\\
Phase 2 -- Solve: Using the equations above, write a Python `solution()` function that computes the answer.
\end{promptbox}

\subsection{AICrypto Prompts}

\subsubsection{Base System Prompt (All Strategies)}

\begin{promptbox}[title=Base System (All Methods)]
\small\ttfamily
You are an expert cryptographer tasked with solving cryptographic proof problems. Your responses must demonstrate deep understanding of cryptographic principles, mathematical rigor, and clear logical reasoning.\\[0.5em]
Output Format Requirements:\\
Your response MUST be structured into exactly two sections:\\[0.3em]
\#\# Analysis\\
- Present your complete thought process\\
- Show all intermediate steps and considerations\\
- This section is for your working/scratch work\\[0.3em]
\#\# Proof\\
- Provide a clean, formal proof suitable for academic submission\\
- This section alone will be graded
\end{promptbox}

\subsubsection{Strategy-Specific System Additions}

\begin{promptbox}[title=Zero-Shot Addition]
\small\ttfamily
(no addition)
\end{promptbox}

\begin{promptbox}[title=CoT Addition]
\small\ttfamily
Approach: Solve problems by thinking step-by-step.
\end{promptbox}

\begin{promptbox}[title=PoT Addition]
\small\ttfamily
Approach: Use algorithmic thinking and construct explicit adversaries/reductions with pseudocode.
\end{promptbox}

\begin{promptbox}[title=F-1 Addition (Ours)]
\small\ttfamily
Approach: Solve problems mainly through equations and formal mathematical derivations.
\end{promptbox}

\subsubsection{User Message Prefixes}

\begin{promptbox}[title=Zero-Shot Prefix]
\small\ttfamily
(none)
\end{promptbox}

\begin{promptbox}[title=CoT Prefix]
\small\ttfamily
Let's think step-by-step.
\end{promptbox}

\begin{promptbox}[title=PoT Prefix]
\small\ttfamily
Construct an explicit algorithm/adversary to solve this.
\end{promptbox}

\begin{promptbox}[title=F-1 Prefix (Ours)]
\small\ttfamily
Solve using equations: identify givens -> write equations -> solve -> verify.
\end{promptbox}

\subsubsection{F-1 Variants (Composability)}

The variants append different system additions and user prefixes onto the shared base system prompt.

\begin{promptbox}[title=F-1+ZS System Addition]
\small\ttfamily
Approach: Use an equation-first method. First formalize the problem with explicit key variables (parameters, oracles, security goals) and governing equations/security definitions, then solve by direct application of the formalization.
\end{promptbox}

\begin{promptbox}[title=F-1+ZS User Prefix]
\small\ttfamily
Phase 1 -- Formalize:\\
1. List the key variables (parameters, oracles, security goals).\\
2. Write the governing equation(s)/security definition(s) using \textbackslash begin\{equation\}...\textbackslash end\{equation\}.\\[0.3em]
Phase 2 -- Solve: Apply the formalization directly to derive the result.
\end{promptbox}

\begin{promptbox}[title=F-1+CoT System Addition]
\small\ttfamily
Approach: Use an equation-first method. First formalize the problem with explicit key variables (parameters, oracles, security goals) and governing equations/security definitions, then derive the result through step-by-step reasoning.
\end{promptbox}

\begin{promptbox}[title=F-1+CoT User Prefix]
\small\ttfamily
Phase 1 -- Formalize: (same as F-1+ZS).\\
Phase 2 -- Solve: Using the formalization above, derive the result through step-by-step reasoning.
\end{promptbox}

\begin{promptbox}[title=F-1+PoT System Addition]
\small\ttfamily
Approach: Use an equation-first method. First formalize the problem with explicit key variables (parameters, oracles, security goals) and governing equations/security definitions, then construct an explicit algorithm/adversary/reduction in pseudocode.
\end{promptbox}

\begin{promptbox}[title=F-1+PoT User Prefix]
\small\ttfamily
Phase 1 -- Formalize: (same as F-1+ZS).\\
Phase 2 -- Solve: Using the formalization above, construct an explicit algorithm/adversary/reduction in pseudocode.
\end{promptbox}

\section{Token Efficiency Analysis}
\label{app:efficiency}

This appendix provides detailed token efficiency analysis across models and benchmarks. We use two complementary metrics following \citet{chen2023frugalgpt} and \citet{lee2025token}:

\paragraph{Metrics Definition}
\begin{itemize}[leftmargin=*]
\setlength\itemsep{-0.1em}
    \item \textbf{Efficiency Ratio} = $\frac{\text{Accuracy}}{\text{Avg Tokens}} \times 100$ (measures accuracy points gained per 100 tokens; higher is better)
    \item \textbf{Tokens per Correct} = $\frac{\text{Avg Tokens}}{\text{Accuracy}/100}$ (measures average tokens needed per correct answer; lower is better)
\end{itemize}

These metrics capture the accuracy-efficiency tradeoff central to practical LLM deployment, where token count determines both latency and operational cost.

\subsection{FinanceMath Token Efficiency by Model}

Table~\ref{tab:eff_finmath} shows token efficiency for FinanceMath (n=200). F-1 achieves the best efficiency ratio for 4 out of 5 models.

\begin{table}[H]
\centering
\small
\resizebox{\columnwidth}{!}{%
\begin{tabular}{@{}lcccc|cccc@{}}
\toprule
 & \multicolumn{4}{c|}{\textbf{Avg Tokens}} & \multicolumn{4}{c}{\textbf{Efficiency Ratio} $\uparrow$} \\
\textbf{Model} & ZS & CoT & PoT & F-1 & ZS & CoT & PoT & F-1 \\
\midrule
GPT-5 & 2,000 & 2,239 & 2,668 & 2,271 & 1.30 & 1.90 & 2.02 & \textbf{2.82} \\
Gemini-2.5-Pro & 2,279 & 2,850 & 2,831 & 2,879 & 1.43 & 1.96 & 1.89 & \textbf{1.95} \\
Qwen3-30B & 2,708 & 3,987 & 3,502 & 3,478 & 1.05 & 1.34 & 1.53 & \textbf{1.61} \\
Qwen3-235B & 2,394 & 2,654 & 3,214 & 3,364 & 1.48 & 1.32 & \textbf{1.85} & 1.81 \\
DeepSeek-V3.1 & 598 & 571 & 711 & 627 & 4.60 & 4.90 & 5.28 & \textbf{7.02} \\
\midrule
\textbf{Average} & 1,996 & 2,460 & 2,585 & 2,524 & 1.97 & 2.29 & 2.51 & \textbf{3.04} \\
\bottomrule
\end{tabular}%
}
\caption{FinanceMath (n=200) token efficiency by model. ZS=Zero-Shot.}
\label{tab:eff_finmath}
\end{table}

\subsection{OlympiadBench Token Efficiency by Model}

Table~\ref{tab:eff_olympiad} shows token efficiency for OlympiadBench (n=1,438), averaged across all subjects.

\begin{table}[H]
\centering
\small
\resizebox{\columnwidth}{!}{%
\begin{tabular}{@{}lcccc|cccc@{}}
\toprule
 & \multicolumn{4}{c|}{\textbf{Avg Tokens}} & \multicolumn{4}{c}{\textbf{Efficiency Ratio} $\uparrow$} \\
\textbf{Model} & ZS & CoT & PoT & F-1 & ZS & CoT & PoT & F-1 \\
\midrule
GPT-5 & 6,742 & 8,593 & 5,947 & 8,204 & 1.27 & 1.12 & 0.96 & \textbf{1.23} \\
Gemini-2.5-Pro & 17,645 & 17,144 & 15,736 & 15,973 & 0.37 & 0.64 & 0.47 & \textbf{0.72} \\
Qwen3-30B & 13,980 & 12,130 & 9,099 & 10,675 & 0.40 & 0.57 & 0.53 & \textbf{0.73} \\
Qwen3-235B & 16,849 & 17,103 & 12,434 & 12,970 & 0.31 & 0.38 & 0.48 & \textbf{0.65} \\
DeepSeek-V3.1 & 2,624 & 2,617 & 2,965 & 2,376 & 2.14 & 1.73 & 1.10 & \textbf{3.71} \\
\midrule
\textbf{Average} & 11,221 & 11,517 & 9,236 & 10,040 & 0.97 & 0.99 & 0.70 & \textbf{1.08} \\
\bottomrule
\end{tabular}%
}
\caption{OlympiadBench (n=1,438) token efficiency by model.}
\label{tab:eff_olympiad}
\end{table}

\subsection{IMO-Bench Token Efficiency by Model}

Table~\ref{tab:eff_imo} shows token efficiency for IMO-Bench (n=460). Note that PoT shows slightly higher efficiency due to shorter outputs, though accuracy remains lower than F-1.

\begin{table}[H]
\centering
\small
\resizebox{\columnwidth}{!}{%
\begin{tabular}{@{}lcccc|cccc@{}}
\toprule
 & \multicolumn{4}{c|}{\textbf{Avg Tokens}} & \multicolumn{4}{c}{\textbf{Efficiency Ratio} $\uparrow$} \\
\textbf{Model} & ZS & CoT & PoT & F-1 & ZS & CoT & PoT & F-1 \\
\midrule
GPT-5 & 17,578 & 17,787 & 14,665 & 17,159 & 0.27 & 0.26 & \textbf{0.30} & 0.27 \\
Gemini-2.5-Pro & 8,885 & 10,023 & 9,495 & 10,308 & 1.09 & 1.09 & \textbf{1.32} & 1.18 \\
Qwen3-30B & 20,986 & 20,112 & 16,012 & 19,794 & 0.19 & 0.21 & 0.21 & \textbf{0.21} \\
Qwen3-235B & 16,119 & 16,124 & 14,931 & 16,086 & 0.14 & 0.13 & \textbf{0.16} & 0.14 \\
DeepSeek-V3.1 & 3,444 & 3,409 & 2,632 & 3,594 & \textbf{0.93} & 0.85 & 0.74 & 0.82 \\
\midrule
\textbf{Average} & 14,670 & 14,715 & 12,421 & 14,540 & 0.48 & 0.47 & \textbf{0.52} & 0.49 \\
\bottomrule
\end{tabular}%
}
\caption{IMO-Bench (n=460) token efficiency by model.}
\label{tab:eff_imo}
\end{table}

\subsection{AICrypto Token Efficiency by Model}

Table~\ref{tab:eff_crypto} shows token efficiency for AICrypto (n=18). GPT-5 data is unavailable for this benchmark.

\begin{table}[H]
\centering
\small
\resizebox{\columnwidth}{!}{%
\begin{tabular}{@{}lcccc|cccc@{}}
\toprule
 & \multicolumn{4}{c|}{\textbf{Avg Tokens}} & \multicolumn{4}{c}{\textbf{Efficiency Ratio} $\uparrow$} \\
\textbf{Model} & ZS & CoT & PoT & F-1 & ZS & CoT & PoT & F-1 \\
\midrule
Gemini-2.5-Pro & 3,367 & 5,105 & 5,000 & 5,288 & 2.42 & 1.82 & 1.84 & \textbf{1.83} \\
Qwen3-30B & 9,768 & 9,466 & 12,092 & 9,800 & 0.76 & 0.70 & 0.64 & \textbf{0.88} \\
Qwen3-235B & 9,909 & 8,570 & 8,360 & 9,282 & 0.66 & \textbf{0.83} & 0.78 & 0.82 \\
DeepSeek-V3.1 & 4,027 & 3,872 & 4,789 & 3,754 & 1.67 & 1.87 & 1.61 & \textbf{2.14} \\
\midrule
\textbf{Average} & 6,768 & 6,753 & 7,560 & 7,031 & 1.38 & 1.31 & 1.22 & \textbf{1.42} \\
\bottomrule
\end{tabular}%
}
\caption{AICrypto (n=18) token efficiency by model. GPT-5 data unavailable.}
\label{tab:eff_crypto}
\end{table}

\subsection{Tokens per Correct Answer Summary}

Table~\ref{tab:tokens_per_correct} summarizes tokens per correct answer across all benchmarks. F-1 requires 23.2\% fewer tokens than Zero-Shot and 11.7\% fewer than CoT on average.

\begin{table}[H]
\centering
\small
\resizebox{\columnwidth}{!}{%
\begin{tabular}{@{}lcccc|c@{}}
\toprule
\textbf{Method} & \textbf{FinMath} & \textbf{Olympiad} & \textbf{IMO} & \textbf{Crypto} & \textbf{Avg} \\
\midrule
Zero-Shot & 6,624 & 51,929 & 51,192 & 9,597 & 29,836 \\
CoT & 5,486 & 37,649 & 51,355 & 9,299 & 25,947 \\
PoT & 4,815 & 37,120 & 43,638 & 9,986 & 23,890 \\
\textbf{F-1 (Ours)} & \textbf{4,368} & \textbf{25,517} & 53,376 & \textbf{8,426} & \textbf{22,922} \\
\bottomrule
\end{tabular}%
}
\caption{Tokens per Correct answer (lower is better).}
\label{tab:tokens_per_correct}
\end{table}

\section{Additional Results by Subtask}
\label{app:results}

This appendix provides detailed results broken down by subtasks within each benchmark. Overall scores are reported in Table~\ref{tab:main_results}.

\subsection{IMO-Bench}

Table~\ref{tab:imo_breakdown} shows results on IMO-Bench subtasks: IMO-AnswerBench (closed-form answers) and IMO-ProofBench (proof-based).

\begin{table}[t]
\centering
\small
\begin{tabular}{@{}lcccc@{}}
\toprule
\textbf{Model} & \textbf{ZS} & \textbf{CoT} & \textbf{PoT} & \textbf{F-1} \\
\midrule
\multicolumn{5}{l}{\textit{IMO-AnswerBench}} \\
\midrule
GPT-5 & 67.75 & \textbf{67.75} & 49.50 & 64.50 \\
Gemini 2.5 Pro & 68.50 & 69.25 & 60.50 & \textbf{69.75} \\
Qwen3-30B & \textbf{61.50} & 60.75 & 54.50 & 61.25 \\
Qwen3-235B & 7.75 & 8.75 & \textbf{15.50} & 7.25 \\
DeepSeek-V3.1 & \textbf{42.75} & 42.50 & 24.00 & 41.00 \\
\midrule
\multicolumn{5}{l}{\textit{IMO-ProofBench}} \\
\midrule
GPT-5 & 44.76 & 41.67 & 45.95 & \textbf{46.67} \\
Gemini 2.5 Pro & 42.38 & 38.57 & 37.62 & \textbf{44.29} \\
Qwen3-30B & 40.48 & \textbf{45.95} & 30.71 & 42.38 \\
Qwen3-235B & 35.95 & 32.14 & 32.38 & \textbf{36.43} \\
DeepSeek-V3.1 & \textbf{18.33} & 12.62 & 10.95 & 14.29 \\
\bottomrule
\end{tabular}
\caption{IMO-Bench breakdown by subtask (\%). ZS = Zero-Shot. Best per row in \textbf{bold}.}
\label{tab:imo_breakdown}
\end{table}

\subsection{OlympiadBench}

Table~\ref{tab:olympiad_full} shows results on OlympiadBench subtasks: OE (open-ended) and TP (theorem-proving) for both mathematics and physics.

\begin{table}[t]
\centering
\small
\begin{tabular}{@{}lcccc@{}}
\toprule
\textbf{Model} & \textbf{ZS} & \textbf{CoT} & \textbf{PoT} & \textbf{F-1} \\
\midrule
\multicolumn{5}{l}{\textit{OE\_maths (n=674)}} \\
\midrule
GPT-5 & \textbf{87.24} & 85.91 & 77.89 & 86.35 \\
Gemini 2.5 Pro & 31.90 & 83.23 & 86.05 & \textbf{86.20} \\
Qwen3-30B & 61.42 & 70.92 & 75.67 & \textbf{78.64} \\
Qwen3-235B & 70.40 & 76.05 & 70.83 & \textbf{77.61} \\
DeepSeek-V3.1 & \textbf{75.07} & 74.63 & 70.18 & 72.11 \\
\midrule
\multicolumn{5}{l}{\textit{OE\_physics (n=236)}} \\
\midrule
GPT-5 & \textbf{46.61} & 42.37 & 40.68 & 44.92 \\
Gemini 2.5 Pro & 55.93 & 53.81 & 50.00 & \textbf{56.36} \\
Qwen3-30B & 40.68 & \textbf{43.22} & 39.41 & 38.14 \\
Qwen3-235B & \textbf{36.60} & 36.32 & 29.66 & \textbf{36.60} \\
DeepSeek-V3.1 & 41.95 & \textbf{48.73} & 39.83 & 44.92 \\
\midrule
\multicolumn{5}{l}{\textit{TP\_maths (n=503)}} \\
\midrule
GPT-5 & 29.22 & 32.21 & 12.33 & \textbf{35.98} \\
Gemini 2.5 Pro & 24.06 & \textbf{25.65} & 13.12 & 24.25 \\
Qwen3-30B & 36.58 & 38.77 & 33.80 & \textbf{41.95} \\
Qwen3-235B & 6.31 & 11.74 & 17.76 & \textbf{24.80} \\
DeepSeek-V3.1 & 7.95 & 10.34 & 5.96 & \textbf{12.13} \\
\midrule
\multicolumn{5}{l}{\textit{TP\_physics (n=25)}} \\
\midrule
GPT-5 & 72.00 & 92.00 & 44.00 & \textbf{96.00} \\
Gemini 2.5 Pro & 56.00 & 68.00 & 36.00 & \textbf{72.00} \\
Qwen3-30B & 68.00 & \textbf{80.00} & 44.00 & \textbf{80.00} \\
Qwen3-235B & \textbf{76.00} & 68.00 & 36.00 & 72.00 \\
DeepSeek-V3.1 & 56.00 & \textbf{76.00} & 20.00 & 72.00 \\
\bottomrule
\end{tabular}
\caption{OlympiadBench breakdown by subtask (\%). OE = Open-Ended, TP = Theorem-Proving. Best per row in \textbf{bold}.}
\label{tab:olympiad_full}
\end{table}

\subsection{FinanceMath}

Table~\ref{tab:finance_breakdown} shows results across FinanceMath categories.

\begin{table}[t]
\centering
\small
\begin{tabular}{@{}lcccc@{}}
\toprule
\textbf{Model} & \textbf{ZS} & \textbf{CoT} & \textbf{PoT} & \textbf{F-1} \\
\midrule
\multicolumn{5}{l}{\textit{Accounting (n=39)}} \\
\midrule
GPT-5 & 25.64 & 58.97 & 61.54 & \textbf{76.92} \\
Gemini 2.5 Pro & 25.64 & 58.97 & 61.54 & \textbf{76.92} \\
Qwen3-30B & 35.90 & 61.54 & 64.10 & \textbf{66.67} \\
Qwen3-235B & 51.28 & 41.03 & 66.67 & \textbf{74.36} \\
DeepSeek-V3.1 & 20.51 & 23.08 & \textbf{58.97} & 43.59 \\
\midrule
\multicolumn{5}{l}{\textit{Asset Classes \& Derivatives (n=63)}} \\
\midrule
GPT-5 & 26.98 & 33.33 & 46.03 & \textbf{63.49} \\
Gemini 2.5 Pro & 28.57 & \textbf{52.38} & 46.03 & \textbf{52.38} \\
Qwen3-30B & 19.05 & \textbf{49.21} & \textbf{49.21} & \textbf{49.21} \\
Qwen3-235B & 22.22 & 26.98 & \textbf{60.32} & 57.14 \\
DeepSeek-V3.1 & 22.22 & 22.22 & 25.40 & \textbf{41.27} \\
\midrule
\multicolumn{5}{l}{\textit{Corporate \& Securities (n=10)}} \\
\midrule
GPT-5 & 20.00 & 40.00 & 60.00 & \textbf{70.00} \\
Gemini 2.5 Pro & 30.00 & \textbf{60.00} & 50.00 & \textbf{60.00} \\
Qwen3-30B & 20.00 & \textbf{60.00} & \textbf{60.00} & \textbf{60.00} \\
Qwen3-235B & 30.00 & 20.00 & \textbf{60.00} & \textbf{60.00} \\
DeepSeek-V3.1 & 10.00 & 20.00 & 20.00 & \textbf{70.00} \\
\midrule
\multicolumn{5}{l}{\textit{Market Analysis \& Economic (n=24)}} \\
\midrule
GPT-5 & 37.50 & 50.00 & 62.50 & \textbf{70.83} \\
Gemini 2.5 Pro & 45.83 & 54.17 & \textbf{66.67} & 54.17 \\
Qwen3-30B & 41.67 & 54.17 & 54.17 & \textbf{62.50} \\
Qwen3-235B & 54.17 & 45.83 & \textbf{62.50} & \textbf{62.50} \\
DeepSeek-V3.1 & 41.67 & 37.50 & \textbf{58.33} & 41.67 \\
\midrule
\multicolumn{5}{l}{\textit{Portfolio Management (n=19)}} \\
\midrule
GPT-5 & 15.79 & 42.11 & \textbf{57.89} & \textbf{57.89} \\
Gemini 2.5 Pro & 36.84 & \textbf{63.16} & 57.89 & \textbf{63.16} \\
Qwen3-30B & 21.05 & 47.37 & \textbf{57.89} & \textbf{57.89} \\
Qwen3-235B & 36.84 & 36.84 & 47.37 & \textbf{68.42} \\
DeepSeek-V3.1 & 31.58 & 26.32 & \textbf{42.11} & 36.84 \\
\midrule
\multicolumn{5}{l}{\textit{Quantitative Analysis (n=36)}} \\
\midrule
GPT-5 & 25.00 & 41.67 & 58.33 & \textbf{61.11} \\
Gemini 2.5 Pro & 38.89 & 55.56 & \textbf{58.33} & 55.56 \\
Qwen3-30B & 36.11 & \textbf{61.11} & 55.56 & 58.33 \\
Qwen3-235B & 33.33 & 41.67 & \textbf{63.89} & 61.11 \\
DeepSeek-V3.1 & 38.89 & 41.67 & 30.56 & \textbf{52.78} \\
\midrule
\multicolumn{5}{l}{\textit{Risk Management (n=9)}} \\
\midrule
GPT-5 & \textbf{22.22} & \textbf{22.22} & \textbf{22.22} & 11.11 \\
Gemini 2.5 Pro & \textbf{22.22} & 11.11 & 11.11 & 11.11 \\
Qwen3-30B & \textbf{22.22} & \textbf{22.22} & 11.11 & \textbf{22.22} \\
Qwen3-235B & \textbf{22.22} & \textbf{22.22} & \textbf{22.22} & 11.11 \\
DeepSeek-V3.1 & \textbf{22.22} & \textbf{22.22} & 11.11 & \textbf{22.22} \\
\bottomrule
\end{tabular}
\caption{FinanceMath breakdown by category (\%). Best per row in \textbf{bold}.}
\label{tab:finance_breakdown}
\end{table}

\subsection{AICrypto}

Table~\ref{tab:crypto_breakdown} shows results across AICrypto categories: Encryptions (ENC), Foundation (FUN), Pseudorandomness (PR), and Signatures (SIGN).

\begin{table}[t]
\centering
\small
\begin{tabular}{@{}lcccc@{}}
\toprule
\textbf{Model} & \textbf{ZS} & \textbf{CoT} & \textbf{PoT} & \textbf{F-1} \\
\midrule
\multicolumn{5}{l}{\textit{Encryptions (n=3)}} \\
\midrule
GPT-5 & 97.90 & \textbf{99.00} & 88.70 & 96.90 \\
Gemini 2.5 Pro & 94.80 & \textbf{100.00} & \textbf{100.00} & \textbf{100.00} \\
Qwen3-30B & 68.00 & 51.50 & \textbf{99.00} & 89.70 \\
Qwen3-235B & 64.40 & \textbf{88.70} & 59.30 & 76.50 \\
DeepSeek-V3.1 & 47.40 & 70.60 & \textbf{79.20} & 78.40 \\
\midrule
\multicolumn{5}{l}{\textit{Foundation (n=5)}} \\
\midrule
GPT-5 & 89.60 & \textbf{98.60} & 83.30 & \textbf{98.60} \\
Gemini 2.5 Pro & 78.40 & 95.70 & 92.80 & \textbf{100.00} \\
Qwen3-30B & 77.70 & 65.50 & 78.80 & \textbf{91.40} \\
Qwen3-235B & 78.40 & 76.60 & 80.60 & \textbf{83.80} \\
DeepSeek-V3.1 & 80.60 & 82.40 & \textbf{84.20} & 78.40 \\
\midrule
\multicolumn{5}{l}{\textit{Pseudorandomness (n=7)}} \\
\midrule
GPT-5 & 78.70 & 97.50 & 95.10 & \textbf{100.00} \\
Gemini 2.5 Pro & 65.60 & 72.10 & 73.80 & \textbf{82.00} \\
Qwen3-30B & 65.60 & 70.50 & \textbf{72.10} & 59.00 \\
Qwen3-235B & 17.20 & 18.00 & 23.00 & \textbf{53.30} \\
DeepSeek-V3.1 & 59.00 & 38.50 & 45.90 & \textbf{77.00} \\
\midrule
\multicolumn{5}{l}{\textit{Signatures (n=3)}} \\
\midrule
GPT-5 & \textbf{100.00} & \textbf{100.00} & \textbf{100.00} & \textbf{100.00} \\
Gemini 2.5 Pro & 84.80 & \textbf{100.00} & \textbf{100.00} & \textbf{100.00} \\
Qwen3-30B & 98.50 & \textbf{100.00} & 21.20 & \textbf{100.00} \\
Qwen3-235B & \textbf{100.00} & 98.50 & \textbf{100.00} & \textbf{100.00} \\
DeepSeek-V3.1 & 84.80 & \textbf{100.00} & 98.50 & 98.50 \\
\bottomrule
\end{tabular}
\caption{AICrypto breakdown by category (\%). Best per row in \textbf{bold}.}
\label{tab:crypto_breakdown}
\end{table}

\section{Qualitative Example Details}
\label{app:examples}

\begin{figure*}[t]
  \centering
  \includegraphics[width=\textwidth]{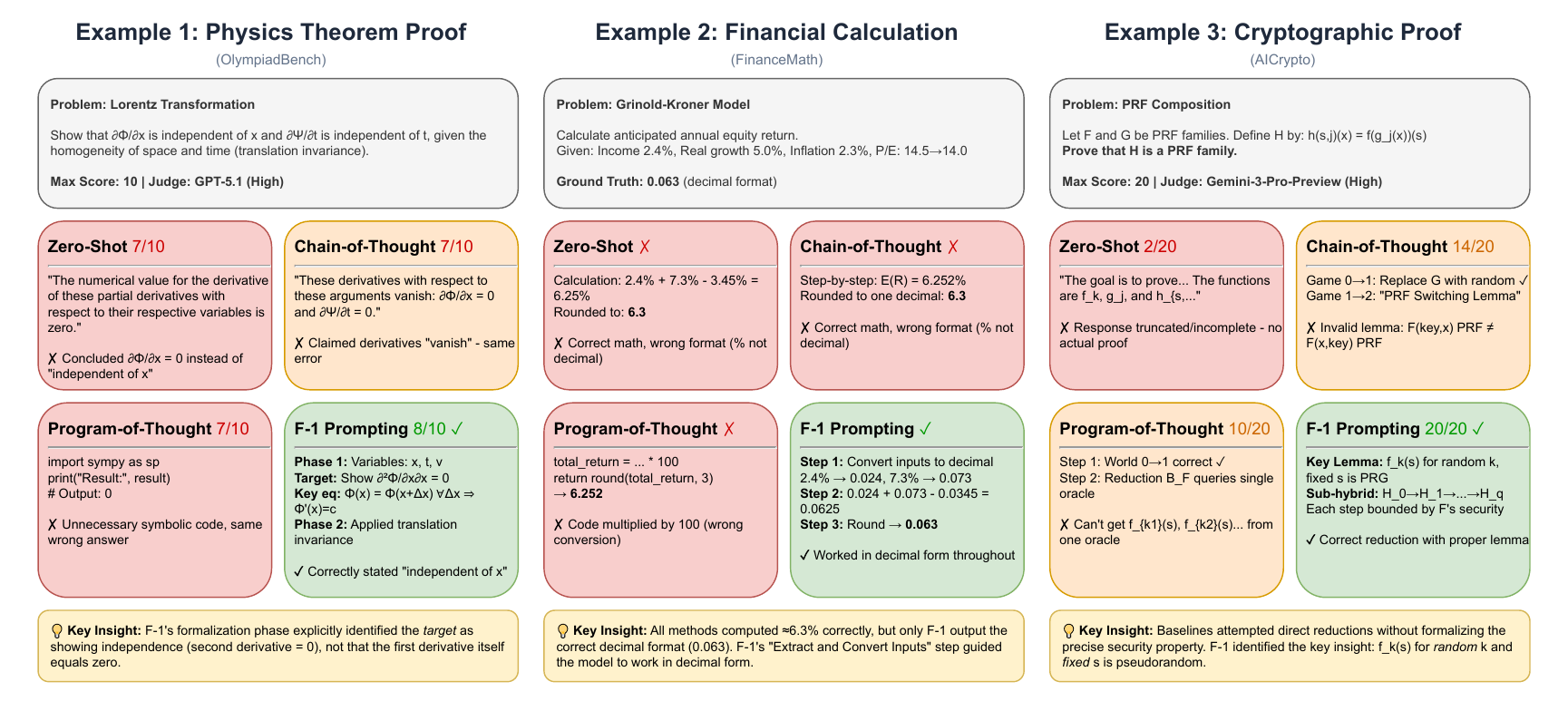}
  \caption{Qualitative comparison. \textbf{(Left) Physics:} Baselines confused ``independent of $x$'' with ``equals zero''; F-1's formalization prevented this. \textbf{(Center) Finance:} Only F-1 output correct decimal format (0.063). \textbf{(Right) Cryptography:} F-1 identified correct sub-hybrid argument while baselines used invalid reductions.}
  \label{fig:qualitative}
\end{figure*}

\Cref{fig:qualitative} shows three representative examples where F-1 succeeds while baselines fail. In \textbf{Physics}, baselines confused ``independent of $x$'' with ``equals zero''; F-1's explicit formalization of the Lorentz equations prevented this semantic confusion. In \textbf{Finance}, all baselines computed the correct value ($\approx$6.3\%) but produced incorrect output formats; F-1's formalization step guided decimal form throughout. In \textbf{Cryptography}, baselines attempted invalid direct reductions, while F-1 identified the correct sub-hybrid argument by first formalizing the PRF security game. These examples demonstrate that F-1's benefits stem from different mechanisms: preventing semantic confusion, enforcing format consistency, and enabling structured proof construction.

This appendix provides full model responses, judge scores, and feedback for the qualitative examples referenced in \Cref{sec:strategy_selection}.

\subsection{Example 1: Lorentz Transformation (OlympiadBench Physics)}

\begin{promptbox}[title=Problem]
\small
Show that $\frac{\partial X}{\partial x}$ is independent of $x$ and $\frac{\partial T}{\partial t}$ is independent of $t$, given the homogeneity of space and time (translation invariance).\\[0.3em]
\textnormal{\textbf{Gold Answer:}} Independence (partial derivatives depend only on $v$, not on $x$ or $t$).
\end{promptbox}

\begin{promptbox}[title=Zero-Shot Response \normalfont{(Judgment: Partial | Score: 7/10)}]
\small
\textbf{By homogeneity (translation invariance):}\\[0.3em]
\textit{Spatial homogeneity:} For any shift $a$, $X(x+a, t, v)$ must relate frames the same way as $X(x, t, v)$, up to the corresponding shift of the primed origin. This implies $X(x+a, t, v) - X(a, 0, v) = X(x, t, v)$. Differentiate with respect to $a$ at $a = 0$ to get $\partial X/\partial x (x, t, v) = \partial X/\partial x (0, 0, v)$, i.e., $\partial X/\partial x$ is independent of $x$. Equivalently, $\partial/\partial x (\partial X/\partial x) = 0$.\\[0.3em]
\textit{Temporal homogeneity:} For any shift $b$, $T(x, t+b, v) - T(0, b, v) = T(x, t, v)$. Differentiate with respect to $b$ at $b = 0$ to get $\partial T/\partial t (x, t, v) = \partial T/\partial t (0, 0, v)$, i.e., $\partial T/\partial t$ is independent of $t$. Equivalently, $\partial/\partial t (\partial T/\partial t) = 0$.\\[0.3em]
Thus, the numerical value for the derivative of these partial derivatives with respect to their respective variables is zero.\\[0.3em]
\texttt{So the final answer is \textbackslash boxed\{0\}.}
\end{promptbox}

\begin{promptbox}[title=Zero-Shot Judge Feedback, colback=yellow!5, colframe=orange!50]
\small
\textnormal{\textbf{Feedback:}} Your use of homogeneity as translation invariance and differentiating with respect to the translation parameter is exactly the right idea and correctly shows that $\partial X/\partial x$ and $\partial T/\partial t$ are independent of $x$ and $t$. The only issue is the final boxed answer: what is constant is these partial derivatives (they can depend on $v$ but not on $x$ or $t$), not that they themselves must equal zero. It would be better to conclude explicitly that $\partial X/\partial x(x,t,v) = \text{constant}(v)$ and $\partial T/\partial t(x,t,v) = \text{constant}(v)$, rather than writing the answer as 0.\\[0.3em]
\textnormal{\textbf{Scores:}} Correctness: 7/10 | Reasoning: 8/10 | Completeness: 7/10
\end{promptbox}

\begin{promptbox}[title=Chain-of-Thought Response \normalfont{(Judgment: Partial | Score: 7/10)}]
\small
\textbf{1) Understand the problem:}\\
We are given general transformations $x' = X(x,t,v)$, $t' = T(x,t,v)$ between inertial frames and asked to use homogeneity of space and time to show that $\partial X/\partial x$ is independent of $x$ and $\partial T/\partial t$ is independent of $t$.\\[0.3em]
\textbf{2) Given and to find:}\\
Given: Homogeneity of space and time (invariance under translations of $x$ and $t$).\\
Need to show: $\partial X/\partial x$ does not depend on $x$; $\partial T/\partial t$ does not depend on $t$.\\[0.3em]
\textbf{3) Break into steps:}\\
Express homogeneity via invariance under shifts: $x \to x + a$, $t \to t + b$. Use this to relate $X(x+a, t+b, v)$ to $X(x,t,v)$ up to a constant. Differentiate to see how partial derivatives behave. Conclude independence.\\[0.3em]
\textbf{4) Solve systematically:}\\
By homogeneity, shifting the origin by $(a, b)$ changes transformed coordinates only by an additive constant:
\begin{align*}
X(x+a, t+b, v) &= X(x, t, v) + C_x(a, b, v)\\
T(x+a, t+b, v) &= T(x, t, v) + C_t(a, b, v)
\end{align*}
Differentiate first equation w.r.t.\ $x$: $\partial X/\partial x (x+a, t+b, v) = \partial X/\partial x (x, t, v)$.\\
Since $a, b$ arbitrary, $\partial X/\partial x$ is invariant under shifts $\Rightarrow$ cannot depend on $x$. Thus $\partial/\partial x (\partial X/\partial x) = 0$.\\
Similarly for $T$: $\partial/\partial t (\partial T/\partial t) = 0$.\\[0.3em]
\textbf{5) Combine results:}\\
The homogeneity implies the $x$-dependence of $\partial X/\partial x$ and $t$-dependence of $\partial T/\partial t$ vanish. Numerically: $\partial/\partial x (\partial X/\partial x) = 0$ and $\partial/\partial t (\partial T/\partial t) = 0$.\\[0.3em]
\texttt{So the final answer is \textbackslash boxed\{0\}.}
\end{promptbox}

\begin{promptbox}[title=CoT Judge Feedback, colback=yellow!5, colframe=orange!50]
\small
\textnormal{\textbf{Feedback:}} You correctly use homogeneity (translation invariance) and the idea of shifting coordinates to argue that the partial derivatives are invariant under shifts, so they cannot depend on $x$ or $t$. The main flaw is in the final statement suggesting the derivatives themselves are zero; what follows from your argument is that their derivatives with respect to $x$ or $t$ vanish, so they are independent of those variables, not that they are numerically zero. Tighten the argument for why translation invariance implies only additive constants, and clarify that the result is independence from $x$ or $t$, not vanishing.\\[0.3em]
\textnormal{\textbf{Scores:}} Correctness: 7/10 | Reasoning: 7/10 | Completeness: 8/10
\end{promptbox}

\begin{promptbox}[title=Program-of-Thought Response \normalfont{(Judgment: Partial | Score: 7/10)}]
\small
\textbf{Step 1 - Reasoning from homogeneity:}\\
Homogeneity (translation invariance) of space and time implies: shifting the origins $x \to x + a$ and $t \to t + b$ should not change the functional form of the boost. Hence for any $a, b$,
\[X(x + a, t + b, v) - X(a, b, v) = X(x, t, v).\]
Differentiate both sides with respect to $a$:
\[\frac{\partial}{\partial a} [X(x + a, t + b, v) - X(a, b, v)] = 0\]
which gives $X_x(x + a, t + b, v) - X_x(a, b, v) = 0$ for all $a, b$.\\
Therefore $X_x$ is independent of $x$ (and $t$), i.e., $\partial X/\partial x$ depends only on $v$.\\
Similarly, $T_t$ is independent of $t$ (and $x$), i.e., $\partial T/\partial t$ depends only on $v$.\\[0.3em]
A practical way to encode this structure is that, due to homogeneity, X and T must be affine in $x$ and $t$:
\begin{align*}
X(x, t, v) &= A(v) x + B(v) t + C(v)\\
T(x, t, v) &= D(v) x + E(v) t + F(v)
\end{align*}
Then $\partial X/\partial x = A(v)$ (independent of $x$), $\partial T/\partial t = E(v)$ (independent of $t$). Equivalently, their second derivatives with respect to $x$ and $t$, respectively, are zero.\\[0.3em]
\textbf{Step 2 - Python code to verify symbolically and extract the numerical result:}\\
We'll use sympy to model the most general affine form implied by homogeneity, and confirm that the $x$-derivative of $\partial X/\partial x$ and the $t$-derivative of $\partial T/\partial t$ are zero. That zero is the requested numerical value.
\begin{lstlisting}[language=Python]
import sympy as sp

# Symbols
x, t, v = sp.symbols('x t v')

# Coefficients can depend on v, but are independent of x and t
A = sp.Function('A')(v)
B = sp.Function('B')(v)
C = sp.Function('C')(v)
D = sp.Function('D')(v)
E = sp.Function('E')(v)
F = sp.Function('F')(v)

# General affine form implied by homogeneity
X = A*x + B*t + C
T = D*x + E*t + F

# First derivatives
dXdx = sp.diff(X, x)    # = A(v), independent of x
dTdt = sp.diff(T, t)    # = E(v), independent of t

# Check independence by differentiating again w.r.t x and t
d2Xdx2 = sp.diff(dXdx, x)  # should be 0
d2Tdt2 = sp.diff(dTdt, t)  # should be 0

print("dX/dx =", dXdx)
print("dT/dt =", dTdt)
print("d^2X/dx^2 =", d2Xdx2)   # numerical value 0
print("d^2T/dt^2 =", d2Tdt2)   # numerical value 0

# For the problem's requested numerical output, we provide the value 0
result = int(d2Xdx2)
print("Result:", result)
\end{lstlisting}
\textbf{Step 3 - Running the code:}
\begin{itemize}[leftmargin=*, nosep]
    \item $\partial X/\partial x = A(v)$
    \item $\partial T/\partial t = E(v)$
    \item $\partial^2 X/\partial x^2 = 0$
    \item $\partial^2 T/\partial t^2 = 0$
    \item Result: 0
\end{itemize}
Therefore, the numerical value that encodes ``independence'' (the derivative of $\partial X/\partial x$ with respect to $x$, and of $\partial T/\partial t$ with respect to $t$) is 0.\\[0.3em]
\texttt{So the final answer is \textbackslash boxed\{0\}.}
\end{promptbox}

\begin{promptbox}[title=PoT Judge Feedback, colback=red!5, colframe=red!50]
\small
\textnormal{\textbf{Feedback:}} You correctly used homogeneity (translation invariance) to derive that $\partial X/\partial x$ and $\partial T/\partial t$ cannot depend on $x$ or $t$, which is exactly the structural point the question targets. However, asserting that X and T must be fully affine in $x$ and $t$ is stronger than what follows strictly from this part of the argument, and the symbolic code plus the reported numerical result 0 are unnecessary and somewhat misleading. What matters is that these derivatives are functions of $v$ only, not that their derivatives are numerically zero. Refine the answer by clearly stating: $\partial X/\partial x = f(v)$ and $\partial T/\partial t = g(v)$, with no additional computation or boxed numerical result.\\[0.3em]
\textnormal{\textbf{Scores:}} Correctness: 7/10 | Reasoning: 7/10 | Completeness: 6/10
\end{promptbox}

\begin{promptbox}[title=F-1 Response \normalfont{(Judgment: Correct | Score: 8/10)}]
\small
\textbf{1) Key variables and target:}\\
Variables: $x, t, v$\\
Target: $\partial X/\partial x$ independence of $x$ and $\partial T/\partial t$ independence of $t$. Numerically equivalent to showing $\partial/\partial x (\partial X/\partial x) = 0$ and $\partial/\partial t (\partial T/\partial t) = 0$.\\[0.3em]
\textbf{2) Main equations:}\\
Definition of partial derivative:
\begin{align*}
\partial X/\partial x(x,t,v) &= \lim_{h \to 0} [X(x+h, t, v) - X(x, t, v)] / h\\
\partial T/\partial t(x,t,v) &= \lim_{k \to 0} [T(x, t+k, v) - T(x, t, v)] / k
\end{align*}
Homogeneity (translation invariance):\\
$X(x+\Delta, t, v) - X(x, t, v)$ depends only on $\Delta$ and $v$ (not on $x$ or $t$).\\
$T(x, t+\Delta, v) - T(x, t, v)$ depends only on $\Delta$ and $v$ (not on $x$ or $t$).\\[0.3em]
\textbf{3) CoT:}\\
By homogeneity, $X(x+h, t, v) - X(x, t, v) = f(h, v)$, independent of $x$ and $t$.\\
Hence $\partial X/\partial x(x,t,v) = \lim_{h\to 0} f(h, v)/h = g(v)$, independent of $x$.\\
Therefore $\partial/\partial x (\partial X/\partial x) = 0$.\\
Similarly, $\partial T/\partial t = r(v)$, so $\partial/\partial t (\partial T/\partial t) = 0$.\\[0.3em]
\textbf{4) Conclusion:}\\
The spatial derivative $\partial X/\partial x$ is independent of $x$; the time derivative $\partial T/\partial t$ is independent of $t$. Numerically: $\partial/\partial x (\partial X/\partial x) = 0$ and $\partial/\partial t (\partial T/\partial t) = 0$.\\[0.3em]
\texttt{So the final answer is \textbackslash boxed\{0\}.}
\end{promptbox}

\begin{promptbox}[title=F-1 Judge Feedback]
\small
\textnormal{\textbf{Feedback:}} Your use of homogeneity via translation invariance and the limit definition of partial derivatives is solid and essentially complete. To improve, focus the conclusion on the functional independence ($\partial X/\partial x = g(v)$, $\partial T/\partial t = r(v)$) rather than framing it as a `numerical' statement with boxed zeros, which can confuse the main conceptual point.\\[0.3em]
\textnormal{\textbf{Scores:}} Correctness: 8/10 | Reasoning: 7/10 | Completeness: 9/10
\end{promptbox}

\begin{promptbox}[title=Error Analysis: Lorentz Transformation, colback=blue!5, colframe=blue!50]
\small
\begin{tabular}{@{}lcccc@{}}
\toprule
\textbf{Method} & \textbf{Judgment} & \textbf{Corr.} & \textbf{Reas.} & \textbf{Comp.} \\
\midrule
Zero-Shot & Partial & 7 & 8 & 7 \\
CoT & Partial & 7 & 7 & 8 \\
PoT & Partial & 7 & 7 & 6 \\
\textbf{F-1} & \textbf{Correct} & \textbf{8} & 7 & \textbf{9} \\
\bottomrule
\end{tabular}\\[0.5em]
\textbf{Key Finding:} F-1's explicit formalization of the \textit{target} (independence) prevented the conceptual error shared by all baselines.
\end{promptbox}

\subsection{Example 2: Market Analysis \& Economics (FinanceMath)}

This example demonstrates a critical error pattern: \textbf{``Correct Calculation, Wrong Format.''} All baseline methods computed the correct numerical value but failed due to output format mismatch (percentage vs.\ decimal).

\begin{promptbox}[title=Problem Statement]
\small
An investor from Australia currently manages an equity portfolio valued at A\$240 million. He is contemplating adjusting the portfolio based on an evaluation of the risk and return possibilities confronting the Australian economy. Using the data in the table, determine the anticipated annual equity return using the Grinold--Kroner model (the number of outstanding shares is presumed to remain unchanged).

\textit{Given data:} Expected income return: 2.4\%, Expected real earnings growth: 5.0\%, Expected inflation: 2.3\%, Current P/E: 14.5x, Expected P/E: 14.0x

\textbf{Ground Truth:} 0.063 (decimal format)
\end{promptbox}

\begin{promptbox}[title=Zero-Shot Response \normalfont{(Judgment: Incorrect | Answer: 6.3)}]
\small
I need to calculate the anticipated annual equity return using the Grinold--Kroner model. The formula is:

Expected Return = Dividend Yield + Earnings Growth + Change in P/E ratio

\textbf{Identify components:}\\
Expected Income Return (dividend yield) = 2.4\%\\
Expected Nominal Earnings Growth = Real growth + Inflation = 5.0\% + 2.3\% = 7.3\%\\
Change in P/E = $(14.0 - 14.5)/14.5 = -0.5/14.5 \approx -3.448\%$\\[0.3em]
\textbf{Calculation:}\\
Total = $2.4\% + 7.3\% + (-3.448\%) = 6.252\%$\\
Rounded to one decimal place (consistent with input data precision): \textbf{6.3\%}\\[0.3em]
\texttt{Final Answer: 6.3}
\end{promptbox}

\begin{promptbox}[title=Chain-of-Thought Response \normalfont{(Judgment: Incorrect | Answer: 6.3)}]
\small
\textbf{Step 1: Identify Grinold--Kroner components}\\
The model decomposes total return into: (1) Income component (dividend yield), (2) Earnings growth component, (3) Change in valuation (P/E ratio)\\[0.3em]
\textbf{Step 2: Extract data and convert}\\
Dividend yield = 2.4\%\\
Nominal earnings growth = 5.0\% (real) + 2.3\% (inflation) = 7.3\%\\
\%$\Delta$(P/E) = $(14.0 - 14.5)/14.5 = -3.448\%$\\[0.3em]
\textbf{Step 3: Apply formula}\\
$E(R) = 2.4\% + 7.3\% - 3.448\% = 6.252\%$\\
Rounded to one decimal: \textbf{6.3\%}\\[0.3em]
\texttt{Final Answer: 6.3}
\end{promptbox}

\begin{promptbox}[title=Program-of-Thought Response \normalfont{(Judgment: Incorrect | Answer: 6.252)}]
\small
\textbf{Reasoning:} Apply Grinold--Kroner model. Since shares outstanding are unchanged, that term is zero.
\begin{lstlisting}[language=Python]
def solution():
    expected_income_return = 2.4
    current_pe = 14.5
    expected_pe = 14.0
    expected_real_earnings_growth = 5.0
    expected_inflation = 2.3

    pe_change_percent = ((expected_pe - current_pe) 
                         / current_pe) * 100
    earnings_growth = expected_real_earnings_growth \
                      + expected_inflation

    total_return = expected_income_return \
                   + pe_change_percent \
                   + earnings_growth
    return round(total_return, 3)

# Result: 6.252
\end{lstlisting}
\texttt{Final Answer: 6.252}
\end{promptbox}

\begin{promptbox}[title=F-1 Response \normalfont{(Judgment: Correct | Answer: 0.063)}]
\small
\textbf{1. Problem Formalization}\\
\textit{Model:} Grinold--Kroner expected equity return\\
\textit{Givens:} Income return = 2.4\%, Real growth = 5.0\%, Inflation = 2.3\%, P/E: $14.5 \to 14.0$\\
\textit{Target:} Expected annual return \textbf{as decimal (not percentage)}\\[0.3em]
\textbf{2. Governing Equation}
\[E(R) = \frac{D}{P} + g_{\text{nominal}} + \frac{\Delta(\text{P/E})}{\text{P/E}_0}\]
\textbf{3. Step-by-Step Calculation (in decimal)}\\
Dividend yield: $0.024$\\
Nominal growth: $0.05 + 0.023 = 0.073$\\
P/E change: $(14.0 - 14.5)/14.5 = -0.034482759$\\
Total: $0.024 + 0.073 - 0.034482759 = 0.062517241$\\[0.3em]
\textbf{4. Format Verification}\\
Round to 3 decimals: $0.062517... \rightarrow \boxed{0.063}$
\end{promptbox}

\begin{promptbox}[title=Error Pattern: ``Correct Calculation{,} Wrong Format'', colback=blue!5, colframe=blue!50]
\scriptsize
\textbf{Observation:} All methods computed $\approx$6.3\% but only F-1 output correct format.

\begin{tabular}{@{}lccp{2.8cm}@{}}
\toprule
\textbf{Method} & \textbf{Computed} & \textbf{Output} & \textbf{Issue} \\
\midrule
Zero-Shot & 6.3\% & 6.3 & \% instead of decimal \\
CoT & 6.3\% & 6.3 & \% instead of decimal \\
PoT & 6.252\% & 6.252 & \texttt{*100} conversion \\
\textbf{F-1} & 6.3\% & \textbf{0.063} & \textit{Correct format} \\
\bottomrule
\end{tabular}

\vspace{0.3em}
\textbf{Root Cause:} Baselines work in percentage throughout without considering output format.\\
\textbf{F-1 Advantage:} Formalization step (``Extract and Convert Inputs'') naturally guides model to work in decimal form, matching expected output format.
\end{promptbox}

\subsection{Example 3: Cryptographic Proof (AICrypto/ProofBench)}

This example demonstrates F-1's advantage in \textbf{cryptographic security proofs}. The problem requires proving that a composed function family is a Pseudorandom Function (PRF), evaluated by an LLM judge on mathematical rigor.

\begin{promptbox}[title=Problem Statement]
\small
Let $F = \{f^n_k: \{0,1\}^n \to \{0,1\}^n\}$ and $G = \{g^n_j: \{0,1\}^n \to \{0,1\}^n\}$ be PRF families. Define the function family $H$ by:
\[h^n_{s,j}(x) = f^n_{g^n_j(x)}(s)\]
where the key is $(s,j) \in \{0,1\}^n \times \{0,1\}^n$. \textbf{Prove that $H$ is a PRF family.}

\textit{Category:} Foundations (FUN) \quad \textit{Max Score:} 20 points \quad \textit{Judge:} Gemini 3 Pro Preview
\end{promptbox}

\begin{promptbox}[title=Zero-Shot Response \normalfont{(Score: 2/20)}]
\small
\textbf{Analysis:}\\
The goal is to prove that $H$ defined by $h_{s,j}(x) = f_{g_j(x)}(s)$ is a PRF family given that $F$ and $G$ are PRF families.

Let's denote the security parameter by $n$. The functions are $f_k$, $g_j$, and $h_{s,j}$...

\textit{[Response truncated/incomplete]}
\end{promptbox}

\begin{promptbox}[title=Zero-Shot Judge Feedback, colback=red!5, colframe=red!50]
\scriptsize
This submission is unfortunately \textbf{incomplete}. You have restated the problem statement and defined the notation, but the proof cuts off immediately after the setup. There is no actual proof, reduction, or analysis provided. I cannot evaluate the mathematical correctness or rigor of an argument that is not present.
\end{promptbox}

\begin{promptbox}[title=Chain-of-Thought Response \normalfont{(Score: 14/20)}]
\small
\textbf{Proof Strategy:} Hybrid argument with two games.\\[0.3em]
\textbf{Game 0 $\to$ Game 1:} Replace $g^n_j$ with random function $\rho$. Reduction to $G$'s security. \checkmark\\[0.3em]
\textbf{Game 1 $\to$ Game 2:} Replace $f^n_{\rho(x)}(s)$ with random function.\\
\textit{Approach:} Use ``PRF Switching Lemma,'' claiming that if $F$ is a PRF, then the dual family $F'$ defined by $f'_s(k) = f_k(s)$ is also a PRF.\\[0.3em]
\textit{[Applies reduction based on this lemma...]}
\end{promptbox}

\begin{promptbox}[title=CoT Judge Feedback, colback=orange!5, colframe=orange!50]
\scriptsize
Your proof structure is logical. However, there is a \textbf{significant flaw} in Game 2 regarding the ``PRF Switching Lemma''. You claim that if $F$ is a PRF, its dual $F'$ (swapping key and input) is also a PRF. \textbf{This is false generally.} A PRF is only guaranteed to be pseudorandom with respect to the input, not the key. For example, if the first bit of the key leaks, $F_k(x)$ might still be a PRF, but $F_x(k)$ would not be. This invalidates the reduction in Game 2.
\end{promptbox}

\begin{promptbox}[title=Program-of-Thought Response \normalfont{(Score: 10/20)}]
\small
\textbf{Step 1 (World 0 $\to$ World 1):} Replace inner PRF $G$ with random function.\\
Reduction to $G$'s security: \textit{correct}. \checkmark\\[0.3em]
\textbf{Step 2 (World 1 $\to$ World 2):} Oracle is $f^n_{\rho(x)}(s)$ where $\rho$ is random.\\
Claim: Distinguishing $f_{k_i}(s)$ (random keys, fixed input) from random relies on PRF security of $F$.\\[0.3em]
\textit{[Constructs reduction $\mathcal{B}_F$ that queries single oracle $O_F$...]}
\end{promptbox}

\begin{promptbox}[title=PoT Judge Feedback, colback=orange!5, colframe=orange!50]
\scriptsize
Step 1 is correct. However, \textbf{Step 2 is incorrect}. PRF security guarantees indistinguishability of $f_k(x_i)$ (fixed key, varying inputs) from random. Here, you have $f_{k_i}(s)$ (varying keys, fixed input).

Your reduction fails: $\mathcal{B}_F$ gets oracle $O_F$ which is either $f_k(\cdot)$ or $R(\cdot)$. It queries $O_F(s)$. But you need $f_{k_1}(s), f_{k_2}(s), \ldots$ for \textit{many different} keys $k_i$. Your reduction only has access to \textit{one} key $k$. You cannot generate samples for multiple keys using a single oracle.
\end{promptbox}

\begin{promptbox}[title=F-1 Response \normalfont{(Score: 20/20)}]
\small
\textbf{1. Goal \& Security Assumptions}\\
Prove $\mathsf{Adv}^{\text{PRF}}_{H,\mathcal{A}}(n)$ is negligible for any PPT $\mathcal{A}$.\\[0.3em]
\textbf{2. Hybrid Argument Structure}\\
\textit{Game 0:} Oracle $O_0(x) = f^n_{g^n_j(x)}(s)$ \quad \textit{Game 1:} Oracle $O_1(x) = f^n_{r_G(x)}(s)$ \quad \textit{Game 2:} Oracle $O_2(x) = r(x)$\\[0.3em]
\textbf{3. Game 0 $\to$ Game 1:} Standard reduction to $G$'s PRF security. $|P_0 - P_1| \le \mathsf{negl}(n)$ \checkmark\\[0.3em]
\textbf{4. Key Lemma for Game 1 $\to$ Game 2:}\\
\textit{``If $F$ is a PRF, then for any fixed $s$, the distribution $\{f^n_k(s)\}_{k \leftarrow \{0,1\}^n}$ is computationally indistinguishable from uniform.''}\\
\textit{Proof:} Build $\mathcal{B}_F$ that queries its oracle at single point $s$. If oracle is $f^n_k$, output is from $\{f^n_k(s)\}$; if random function, output is uniform.\\[0.3em]
\textbf{5. Sub-Hybrid Argument:} Define $H_0, \ldots, H_q$ where $H_i$ answers first $i$ queries with random values. Each $|P(H_{i-1}) - P(H_i)|$ bounded by $F$'s security. Sum over $q$ steps remains negligible.\\[0.3em]
\textbf{Conclusion:} $|P_0 - P_2| \le |P_0 - P_1| + |P_1 - P_2| \le \mathsf{negl}(n)$ \quad $\therefore H$ is a PRF family. $\square$
\end{promptbox}

\begin{promptbox}[title=F-1 Judge Feedback]
\scriptsize
This is an \textbf{excellent, textbook-quality proof}. The hybrid argument is clearly structured, and the reduction strategies for both steps are handled correctly. You correctly identified that replacing $G$ with a random function results in independent keys for $F$. The ``swap'' logic is implicitly handled correctly by noting that we rely on the pseudorandomness of $F$ with respect to a random key, even though the input $s$ is fixed. Your sub-hybrid argument for the $F$-transition is rigorous and correctly bounds the advantage. \textbf{Well done.}
\end{promptbox}

\begin{promptbox}[title=Error Pattern: Incorrect Security Reductions, colback=blue!5, colframe=blue!50]
\scriptsize
\textbf{Observation:} All baselines failed to correctly handle the ``varying keys, fixed input'' scenario.

\begin{tabular}{@{}lcp{4.2cm}@{}}
\toprule
\textbf{Method} & \textbf{Score} & \textbf{Critical Flaw} \\
\midrule
Zero-Shot & 2/20 & Response incomplete (truncated) \\
CoT & 14/20 & Invalid ``PRF Switching Lemma'' assumption \\
PoT & 10/20 & Reduction uses single oracle for multiple keys \\
\textbf{F-1} & \textbf{20/20} & \textit{Correct sub-hybrid with key lemma} \\
\bottomrule
\end{tabular}

\vspace{0.3em}
\textbf{Root Cause:} Baselines attempted direct reductions without formalizing the precise security property needed.\\
\textbf{F-1 Advantage:} Formalization identified the key insight: $f^n_k(s)$ for \textit{random} $k$ and \textit{fixed} $s$ is pseudorandom, enabling a valid sub-hybrid argument.
\end{promptbox}

\section{Evaluation Prompts}
\label{app:eval_prompts}

This appendix provides the complete evaluation prompts used for each benchmark. We use two evaluation paradigms: (1) \textbf{rule-based scoring} with deterministic regex-based extraction for benchmarks with definitive numerical answers, and (2) \textbf{LLM-as-Judge} for proof-based benchmarks requiring semantic evaluation.

\subsection{Rule-Based Scoring Benchmarks}

\paragraph{Code Execution}
For PoT baselines and F-1's PoT strategy, generated code is executed in a sandboxed environment (30s timeout, standard libraries). When F-1 selects PoT as its solving strategy, the generated code is executed identically to the baseline, with the output used as the final answer.

\paragraph{FinanceMath \& OlympiadBench}
These benchmarks have definitive numerical answers. We extract the final answer using regex patterns matching \texttt{\textbackslash boxed\{\}} or the last numerical value, then compare against ground truth with tolerance $\epsilon = 10^{-6}$ for floating-point answers.

\paragraph{IMO-AnswerBench}
For IMO problems with definitive answers, we use the deterministic autograder from \citet{luong-etal-2025-towards}:

\begin{promptbox}[title=IMO-AnswerBench Auto-Grader Prompt, colback=gray!5, colframe=gray!60]
\scriptsize
\textbf{System Role:} Deterministic Mathematical Autograder

You are a precise, automated grading system. Your sole function is to determine if the final answer provided in the Model Solution is mathematically equivalent to the Golden Answer. You must NOT grade the reasoning or steps, only the final result.

\textbf{Equivalence Rules:}
\begin{itemize}[leftmargin=*, nosep]
\item \textbf{Algebraic:} e.g., `n(n+1)/2' $\equiv$ `n\^{}2/2 + n/2'
\item \textbf{Numerical:} e.g., `1/2' $\equiv$ `0.5'; `sqrt(2)/2' $\equiv$ `1/sqrt(2)'
\item \textbf{Set/List:} Order does not matter unless specified as ordered tuple
\item \textbf{No Partial Credit:} Incomplete or partially incorrect $\rightarrow$ Incorrect
\end{itemize}

\textbf{Output:} \textbackslash boxed\{Correct\} or \textbackslash boxed\{Incorrect\}
\end{promptbox}

\subsection{LLM-as-Judge Benchmarks}

For proof-based benchmarks where correctness requires semantic understanding, we employ LLM-as-Judge evaluation with the best-performing models available for each domain: Gemini-3-Pro (High) for IMO-ProofBench and AICrypto, GPT-5.1 (High) for OlympiadBench TP.

\paragraph{IMO-ProofBench}
Mathematical proofs are evaluated on a 0--7 scale following \citet{luong-etal-2025-towards}:

\begin{promptbox}[title=IMO-ProofBench Judge Prompt, colback=blue!5, colframe=blue!50]
\scriptsize
You are an expert grader for the International Mathematics Olympiad (IMO). Evaluate the proposed solution strictly and rigorously.

\textbf{Scoring Rubric (0--7 scale):}
\begin{itemize}[leftmargin=*, nosep]
\item \textbf{7 Points (Correct):} Complete, correct, and fully rigorous
\item \textbf{6 Points (Almost):} Sound core argument with minor errors/gaps
\item \textbf{1 Point (Partial):} Substantial progress on key steps
\item \textbf{0 Points (Incorrect):} No substantial progress or fundamentally flawed
\end{itemize}

\textbf{Evaluation Process:}
\begin{enumerate}[leftmargin=*, nosep]
\item Analyze problem and ground truth solution
\item Step-by-step verification of every logical step
\item Identify all flaws, gaps, and errors
\item Compare against grading guidelines
\end{enumerate}

\textbf{Output:} \texttt{<points>N out of 7</points>}
\end{promptbox}

\paragraph{AICrypto (CryptoProof)}
Cryptographic proofs are evaluated on correctness, completeness, and rigor:

\begin{promptbox}[title=AICrypto Judge Prompt]
\scriptsize
\textbf{System:} You are a professor of theoretical computer science specializing in mathematical foundations of cryptography, grading student homework for CS 6857 (Graduate Cryptography).

\textbf{Grading Rubric (100 points, scaled to problem max):}
\begin{itemize}[leftmargin=*, nosep]
\item \textbf{Mathematical Correctness (60\%):} Sound reasoning, justified claims, correct reductions, accurate probability analysis
\item \textbf{Completeness (25\%):} All cases covered, no logical gaps, edge cases handled
\item \textbf{Rigor (15\%):} Formal and precise, correct use of definitions, consistent notation
\end{itemize}

\textbf{Common Issues:}
\begin{enumerate}[leftmargin=*, nosep]
\item Incorrect use of cryptographic definitions (confusing OWF with PRF)
\item Missing verification of critical conditions
\item Intuitive arguments without formal justification
\item Logical gaps in reduction proofs
\item Incorrect probability analysis
\end{enumerate}

\textbf{Output:} JSON with score, feedback, key\_errors, strengths
\end{promptbox}

\subsection{Evaluation Summary}

Table~\ref{tab:eval_summary} summarizes the evaluation methodology for each benchmark.

\begin{table}[H]
\centering
\small
\resizebox{\columnwidth}{!}{%
\begin{tabular}{@{}llll@{}}
\toprule
\textbf{Benchmark} & \textbf{Method} & \textbf{Scale} & \textbf{Threshold} \\
\midrule
FinanceMath & Rule-Based (regex) & Binary & Exact match \\
OlympiadBench OE & Rule-Based (regex) & Binary & $\epsilon = 10^{-6}$ \\
OlympiadBench TP & LLM Judge & Binary & Correct/Incorrect \\
IMO-AnswerBench & Rule-Based (LLM) & Binary & Equivalent \\
IMO-ProofBench & LLM Judge & 0--7 & $\geq 6$ = Correct \\
AICrypto & LLM Judge & 0--100 & $\geq 80\%$ = Correct \\
\bottomrule
\end{tabular}%
}
\caption{Evaluation methodology summary. LLM Judge uses Gemini-3-Pro (High) for IMO-ProofBench and AICrypto, GPT-5.1 (High) for OlympiadBench TP.}
\label{tab:eval_summary}
\end{table}

\section{Variant Accuracy and Statistical Tests}
\label{app:variant_accuracy}

\begin{figure}[H]
  \centering
  \includegraphics[width=\columnwidth]{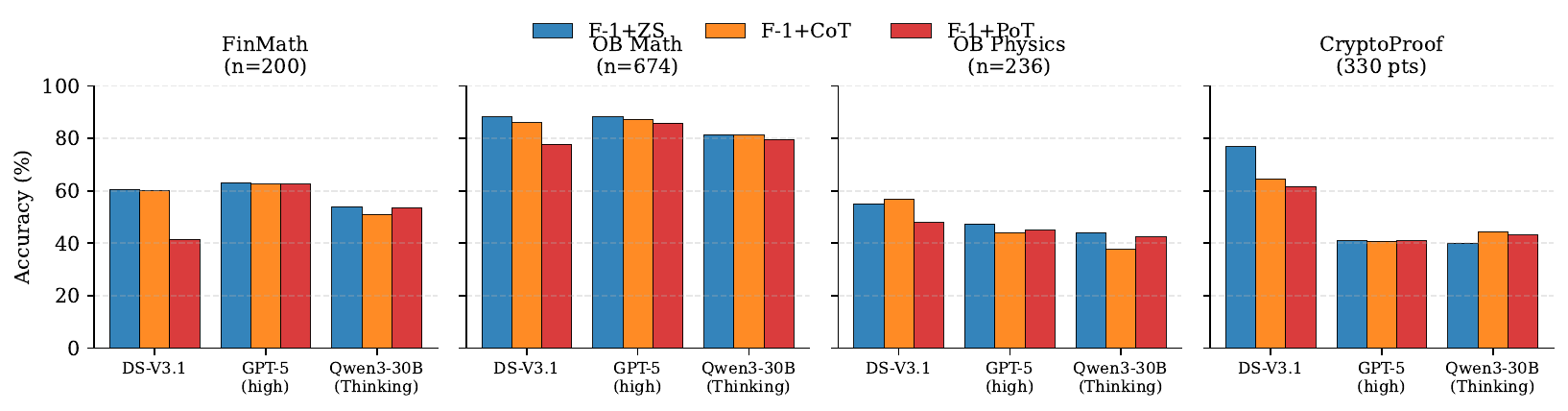}
  \caption{Visual comparison of F-1 variant accuracy at matched maximum reasoning effort. Same data as \Cref{tab:variant_accuracy}.}
  \label{fig:variant_accuracy}
\end{figure}

\begin{table}[H]
\centering
\small
\setlength{\tabcolsep}{4pt}
\resizebox{\columnwidth}{!}{%
\begin{tabular}{@{}llcccc@{}}
\toprule
\textbf{Model} & \textbf{Variant} & \textbf{FinMath} & \textbf{OB Math} & \textbf{OB Phys} & \textbf{Crypto} \\
\midrule
\multirow{3}{*}{DeepSeek-V3.1}
 & F-1+ZS  & \textbf{60.50} & \textbf{88.13} & 55.08          & \textbf{76.8} \\
 & F-1+CoT & 60.00          & 86.05          & \textbf{56.78} & 64.5          \\
 & F-1+PoT & 41.50          & 77.74          & 47.88          & 61.6          \\
\midrule
\multirow{3}{*}{GPT-5 (high)}
 & F-1+ZS  & \textbf{63.00} & \textbf{88.13} & \textbf{47.46} & \textbf{40.9} \\
 & F-1+CoT & 62.50          & 87.24          & 44.07          & 40.8          \\
 & F-1+PoT & 62.50          & 85.76          & 44.92          & \textbf{40.9} \\
\midrule
\multirow{3}{*}{Qwen3-30B-Thk}
 & F-1+ZS  & \textbf{54.00} & \textbf{81.45} & \textbf{44.07} & 40.0          \\
 & F-1+CoT & 51.00          & 81.31          & 37.71          & \textbf{44.5} \\
 & F-1+PoT & 53.50          & 79.67          & 42.37          & 43.1          \\
\bottomrule
\end{tabular}%
}
\caption{Accuracy (\%) across F-1 variants at matched maximum reasoning effort. F-1+ZS is the strongest or tied-strongest variant on 9 of 12 (model$\times$benchmark) cells. F-1+PoT is the weakest variant on math-heavy benchmarks for DeepSeek-V3.1 and GPT-5.}
\label{tab:variant_accuracy}
\end{table}

\Cref{tab:variant_stats} reports pairwise statistical comparisons across the three F-1 variants on OlympiadBench OE benchmarks for which per-problem correctness is recorded.

\begin{table}[H]
\centering
\footnotesize
\setlength{\tabcolsep}{3pt}
\resizebox{\columnwidth}{!}{%
\begin{tabular}{@{}llrrr@{}}
\toprule
\textbf{Bench / Comparison} & \textbf{Model} & \textbf{$\Delta$\%} & \textbf{95\% CI} & \textbf{p} \\
\midrule
\multicolumn{5}{@{}l}{\textit{OB Math (n=674)}} \\
F-1+ZS vs F-1+PoT  & DeepSeek & $+$10.39 & [$+$7.42,\,$+$13.50] & \textbf{$<$0.001} \\
F-1+CoT vs F-1+PoT & DeepSeek &  $+$8.31 & [$+$5.49,\,$+$11.28] & \textbf{$<$0.001} \\
F-1+ZS vs F-1+CoT  & DeepSeek &  $+$2.08 & [$-$0.15,\,$+$4.30]  & 0.087             \\
F-1+ZS vs F-1+PoT  & GPT-5    &  $+$2.37 & [$+$0.30,\,$+$4.60]  & \textbf{0.044}    \\
F-1+ZS vs F-1+PoT  & Qwen3    &  $+$1.78 & [$-$0.45,\,$+$4.01]  & 0.162             \\
\midrule
\multicolumn{5}{@{}l}{\textit{OB Physics (n=236)}} \\
F-1+ZS vs F-1+PoT  & DeepSeek &  $+$7.20 & [$+$1.27,\,$+$13.14] & \textbf{0.024}    \\
F-1+CoT vs F-1+PoT & DeepSeek &  $+$8.90 & [$+$2.97,\,$+$14.83] & \textbf{0.005}    \\
F-1+ZS vs F-1+CoT  & Qwen3    &  $+$6.36 & [$+$1.27,\,$+$11.02] & \textbf{0.020}    \\
F-1+ZS vs F-1+CoT  & GPT-5    &  $+$3.39 & [$-$0.85,\,$+$7.63]  & 0.185             \\
\bottomrule
\end{tabular}%
}
\caption{Pairwise statistical tests on OlympiadBench OE for F-1 variants. McNemar's exact binomial $p$-values; bootstrap 95\% CIs from 10{,}000 resamples. Bold marks $p<0.05$.}
\label{tab:variant_stats}
\end{table}

\section{Strategy Distribution Analysis}
\label{app:strategy_distribution}

\begin{figure}[H]
  \centering
  \includegraphics[width=\columnwidth]{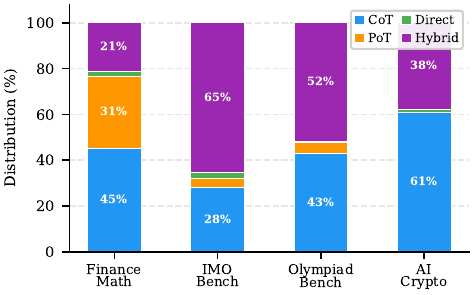}
  \caption{F-1 strategy distribution across domains (pooled over 5 models). Without explicit routing, F-1 naturally favors PoT for computation-heavy FinanceMath, Hybrid for competition math (IMO-Bench), and CoT for proof-oriented AICrypto.}
  \label{fig:strategy_dist}
\end{figure}

To understand \textit{how} F-1 adapts its solving behavior across domains, we classify the strategy used in each F-1 response based on output characteristics (presence of code blocks, step markers, and formula density). \Cref{tab:strategy_dist} and \Cref{fig:strategy_dist} present the distribution across benchmarks, averaged over all models.

\begin{table}[H]
\centering
\small
\resizebox{\columnwidth}{!}{%
\begin{tabular}{@{}lrccccr@{}}
\toprule
\textbf{Benchmark} & \textbf{N} & \textbf{CoT} & \textbf{PoT} & \textbf{Direct} & \textbf{Hybrid} & \textbf{F-ratio} \\
\midrule
FinanceMath & 1,000 & 45.3 & 31.4 & 2.3 & 21.0 & 28.2 \\
IMO-Bench & 2,300 & 28.0 & 4.3 & 2.4 & 65.3 & 29.8 \\
OlympiadBench & 5,752 & 43.0 & 4.9 & 0.5 & 51.6 & 26.5 \\
AICrypto & 90 & 61.1 & 0.0 & 1.1 & 37.8 & 15.9 \\
\bottomrule
\end{tabular}%
}
\caption{F-1 strategy distribution (\%) across benchmarks, pooled over 5 models. \textbf{F-ratio}: average formula density in responses. F-1 strongly favors PoT on computation-heavy FinanceMath (31.4\%) while selecting CoT/Hybrid for proof-oriented AICrypto and IMO-Bench. Per-model breakdowns are in \Cref{app:strategy_per_model}.}
\label{tab:strategy_dist}
\end{table}

The distribution reveals clear domain adaptation: FinanceMath has the highest PoT usage (31.4\%), reflecting numerical computation demands; AICrypto uses exclusively CoT and Hybrid strategies (0\% PoT), consistent with proof problems; and IMO-Bench shows the highest Hybrid usage (65.3\%). Notably, these distributions emerge without explicit routing rules, reflecting each model's internal assessment of equation structure.

\section{Per-Model Strategy Distribution}
\label{app:strategy_per_model}

\Cref{tab:strategy_per_model} provides per-model strategy distributions for F-1 across all benchmarks.

\begin{table}[H]
\centering
\small
\resizebox{\columnwidth}{!}{%
\begin{tabular}{@{}llrrrr@{}}
\toprule
\textbf{Benchmark} & \textbf{Model} & \textbf{CoT} & \textbf{PoT} & \textbf{Direct} & \textbf{Hybrid} \\
\midrule
\multirow{5}{*}{FinanceMath}
 & GPT-5 & 0.0 & 92.0 & 6.0 & 2.0 \\
 & Gemini 2.5 Pro & 28.0 & 44.5 & 0.0 & 27.5 \\
 & Qwen3-30B & 69.5 & 0.0 & 0.0 & 30.5 \\
 & Qwen3-235B & 78.0 & 0.0 & 2.0 & 20.0 \\
 & DeepSeek-V3.1 & 51.0 & 20.5 & 3.5 & 25.0 \\
\midrule
\multirow{5}{*}{IMO-Bench}
 & GPT-5 & 36.1 & 21.1 & 10.4 & 32.4 \\
 & Gemini 2.5 Pro & 18.5 & 0.2 & 1.7 & 79.6 \\
 & Qwen3-30B & 19.3 & 0.0 & 0.0 & 80.7 \\
 & Qwen3-235B & 20.0 & 0.0 & 0.0 & 80.0 \\
 & DeepSeek-V3.1 & 45.9 & 0.0 & 0.0 & 54.1 \\
\midrule
\multirow{4}{*}{OlympiadBench}
 & GPT-5 & 36.1 & 15.4 & 0.8 & 47.8 \\
 & Gemini 2.5 Pro & 33.0 & 1.5 & 1.2 & 64.3 \\
 & Qwen3-30B & 47.6 & 0.0 & 0.0 & 52.4 \\
 & DeepSeek-V3.1 & 55.2 & 2.8 & 0.0 & 42.0 \\
\midrule
\multirow{5}{*}{AICrypto}
 & GPT-5 & 72.2 & 0.0 & 5.6 & 22.2 \\
 & Gemini 2.5 Pro & 33.3 & 0.0 & 0.0 & 66.7 \\
 & Qwen3-30B & 66.7 & 0.0 & 0.0 & 33.3 \\
 & Qwen3-235B & 55.6 & 0.0 & 0.0 & 44.4 \\
 & DeepSeek-V3.1 & 77.8 & 0.0 & 0.0 & 22.2 \\
\bottomrule
\end{tabular}%
}
\caption{Per-model F-1 strategy distribution (\%). GPT-5 heavily favors PoT on FinanceMath (92\%) while Qwen models prefer CoT/Hybrid. On IMO-Bench, Qwen and Gemini models predominantly use Hybrid ($\sim$80\%), while GPT-5 distributes more evenly. Note: OlympiadBench is missing Qwen3-235B data.}
\label{tab:strategy_per_model}
\end{table}

\section{Efficiency, Upper Bound, and Win/Tie/Loss}
\label{app:upper_bound_wtl}

\begin{table}[H]
\centering
\small
\resizebox{\columnwidth}{!}{%
\begin{tabular}{@{}lrrrrr@{}}
\toprule
\textbf{Benchmark} & \textbf{All$\times$} & \textbf{All$\checkmark$} & \textbf{Adapt$\checkmark$} & \textbf{Adapt$\times$} & \textbf{F-1 Only} \\
\midrule
IMO-Bench & 90.7 & 4.8 & 2.5 & 1.8 & 0.3 \\
OlympiadBench & 33.5 & 31.5 & 21.3 & 10.5 & 3.1 \\
FinanceMath & 33.0 & 19.6 & 29.1 & 12.8 & 5.5 \\
AICrypto$^\dagger$ & 18.9 & 37.8 & 21.1 & 14.4 & 7.8 \\
\bottomrule
\end{tabular}%
}
\caption{Per-problem outcome analysis (\%, averaged across 5 models). \textbf{All$\times$}/\textbf{All$\checkmark$}: all four methods fail/succeed (uninformative for strategy choice). \textbf{Adapt$\checkmark$}: F-1 and at least one baseline succeed (F-1 picks a working strategy). \textbf{Adapt$\times$}: at least one baseline succeeds but F-1 fails. \textbf{F-1 Only}: only F-1 succeeds. $^\dagger$$n{=}18$, interpret with caution.}
\label{tab:strategy_analysis}
\end{table}

\begin{table}[H]
\centering
\small
\resizebox{\columnwidth}{!}{%
\begin{tabular}{@{}lcccc|cc@{}}
\toprule
 & \multicolumn{4}{c|}{\textbf{Efficiency Ratio} $\uparrow$} & & \\
\textbf{Method} & FinMath & Olympiad & IMO & Crypto & \textbf{Avg} & \textbf{Overhead} \\
\midrule
Zero-Shot & 1.97 & 0.97 & 0.48 & 1.38 & 1.20 & --- \\
CoT & 2.29 & 0.99 & 0.47 & 1.31 & 1.26 & +35 tok \\
PoT & 2.51 & 0.70 & \textbf{0.52} & 1.22 & 1.24 & +64 tok \\
\textbf{F-1 (Ours)} & \textbf{3.04} & \textbf{1.08} & 0.49 & \textbf{1.42} & \textbf{1.51} & +68 tok \\
\bottomrule
\end{tabular}%
}
\caption{Token efficiency (Efficiency Ratio = Accuracy/Tokens $\times$ 100) and prompt overhead vs.\ Zero-Shot. F-1 achieves the highest efficiency with minimal overhead (+68 tokens). All methods use a single API call.}
\label{tab:efficiency}
\end{table}

\begin{table}[H]
\centering
\small
\begin{tabular}{@{}lccc@{}}
\toprule
\textbf{Benchmark} & \textbf{Upper Bound} & \textbf{F-1} & \textbf{F-1 / UB} \\
\midrule
FinanceMath & 67.0\% & 54.2\% & 80.9\% \\
OlympiadBench & 66.5\% & 55.9\% & 84.1\% \\
AICrypto$^\dagger$ & 81.1\% & 66.7\% & 82.2\% \\
\bottomrule
\end{tabular}
\caption{Upper bound (UB) on applied domains, following analysis in Self-Consistency \citep{wang2023selfconsistency} and ToT \citep{yao2023tree}. Upper Bound = $100\% - \text{All Failed}$ represents the best achievable by \textit{any single baseline}. F-1 = All$\checkmark$ + Adapt$\checkmark$ + F-1 Only (from \Cref{tab:strategy_analysis}). $^\dagger$n=18, interpret with caution.}
\label{tab:upper_bound}
\end{table}

\begin{table}[H]
\centering
\small
\begin{tabular}{@{}lccc|ccc|ccc@{}}
\toprule
 & \multicolumn{3}{c|}{\textbf{vs Zero-Shot}} & \multicolumn{3}{c|}{\textbf{vs CoT}} & \multicolumn{3}{c}{\textbf{vs PoT}} \\
\textbf{Model} & W & T & L & W & T & L & W & T & L \\
\midrule
GPT-5 & 3 & 0 & 1 & 3 & 0 & 1 & 4 & 0 & 0 \\
Gemini & 4 & 0 & 0 & 4 & 0 & 0 & 4 & 0 & 0 \\
Qwen3-30B & 4 & 0 & 0 & 3 & 0 & 1 & 4 & 0 & 0 \\
Qwen3-235B & 3 & 0 & 1 & 4 & 0 & 0 & 3 & 0 & 1 \\
DeepSeek & 3 & 0 & 1 & 3 & 0 & 1 & 4 & 0 & 0 \\
\midrule
\textbf{Total} & 17 & 0 & 3 & 17 & 0 & 3 & 19 & 0 & 1 \\
\bottomrule
\end{tabular}
\caption{Win/Tie/Loss counts for F-1 vs each baseline across 4 benchmarks per model (20 comparisons per baseline). F-1 wins 53/60 total comparisons (88.3\%).}
\label{tab:wtl}
\end{table}

\section{Weighted Average Results}

\Cref{tab:main_results} reports macro-averaged accuracy (unweighted by dataset size) to prevent larger benchmarks from dominating. For completeness, \Cref{tab:weighted_avg} provides weighted averages, where each benchmark contributes proportionally to its size (IMO-Bench: 460, OlympiadBench: 1,438, FinanceMath: 200, AICrypto: 18).

\begin{table}[H]
\centering
\small
\begin{tabular}{@{}lcc@{}}
\toprule
\textbf{Method} & \textbf{Macro Avg} & \textbf{Weighted Avg} \\
\midrule
Zero-Shot & 49.50 & 46.13 \\
CoT & 55.30 & 51.86 \\
PoT & 52.64 & 42.11 \\
\textbf{F-1 (Ours)} & \textbf{61.07} & \textbf{54.54} \\
\bottomrule
\end{tabular}
\caption{Macro vs weighted average accuracy (\%). F-1 leads in both metrics. The gap between macro and weighted is smaller for F-1, indicating consistent performance across benchmark sizes.}
\label{tab:weighted_avg}
\end{table}

\section{Pretraining Pattern Analysis: Supplementary Figures}
\label{app:infini-gram}

\begin{figure}[H]
  \centering
  \includegraphics[width=\columnwidth]{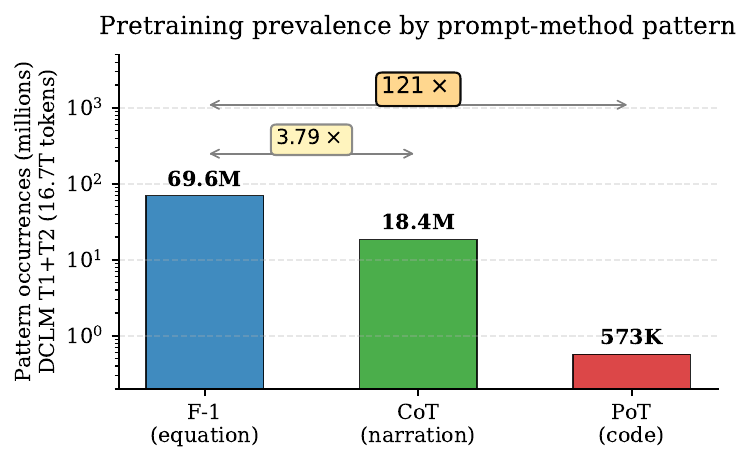}
  \caption{Pretraining prevalence by prompt-method pattern (DCLM-baseline T1+T2 occurrences, log scale). Regenerated from raw infini-gram counts. Equation language (F-1) appears 3.79$\times$ as often as step-by-step narration (CoT) and 121$\times$ as often as code (PoT), a distributional pattern consistent with the observed F-1 vs.\ PoT gap on applied math (correlational, not causal).}
  \label{fig:eq-vs-code}
\end{figure}

This appendix provides the full infini-gram-mini analysis supporting \Cref{sec:infinigram}. We query eight pretraining corpora ($\sim$81.7T tokens): DCLM-baseline (16.7T) and seven Common Crawl snapshots (CC-2025-05 through CC-2025-30). Patterns are classified into three tiers: T1 prompt-derived, T2 shared math/reasoning language (including LaTeX notation), and T3 broad proxies. \Cref{fig:eq-vs-code} summarizes the DCLM findings.

\paragraph{Pattern lists.} \Cref{tab:patterns} enumerates the representative patterns queried per method, grouped by tier (T1 prompt-derived, T2 shared math/code language, T3 broad proxies).

\begin{table}[H]
\centering
\footnotesize
\setlength{\tabcolsep}{3pt}
\begin{tabular}{@{}llp{5.0cm}@{}}
\toprule
\textbf{Method} & \textbf{Tier} & \textbf{Representative patterns} \\
\midrule
\multirow{3}{*}{CoT}
 & T1 & ``step by step'', ``step-by-step'', ``think step by step'', ``solve step by step'' \\
 & T2 & ``first, we need to'', ``let's start by'', ``let's break this down'', ``therefore, the answer is'' \\
 & T3 & ``next, we'' \\
\midrule
\multirow{3}{*}{PoT}
 & T1 & ``write code to'', ``write a program to'', ``write Python code to solve'' \\
 & T2 & \verb|def solve(|, \verb|def solution(|, \verb|return answer|, \verb|print(answer)| \\
 & T3 & \verb|import numpy|, \verb|import math|, \verb|import sympy| \\
\midrule
\multirow{4}{*}{FP / F-1}
 & T1 narrative & ``governing equation'', ``key variables'', ``key equations'', ``identify givens and target'' \\
 & T1 \LaTeX{} & \verb|\begin{equation}|, \verb|\begin{align}|, plus starred and end variants \\
 & T2 narrative & ``solve for'', ``the formula for'', ``the equation is'', ``using the formula'', ``we need to find'' \\
 & T2 \LaTeX{} & \verb|\frac{|, \verb|\sqrt{|, \verb|\sum_{|, \verb|\int_{|, \verb|\text{|, \verb|\cdot|, \verb|\leq|, \verb|\geq|, \verb|\implies|, \verb|\equiv| \\
\bottomrule
\end{tabular}
\caption{Method-associated patterns queried in \Cref{sec:infinigram}. Tier assignment validated empirically via n-gram analysis of actual model outputs: T1 ratios exceed 10$\times$ between own-method and other-method outputs; T2 ratios fall in 0.7--1.7$\times$; T3 ratios drop below 1$\times$. The FP/F-1 tiers split into \textit{narrative} (English phrases from the prompt) and \LaTeX{} (markup tokens the prompt asks the model to emit).}
\label{tab:patterns}
\end{table}

\paragraph{LaTeX pattern inclusion.} The F-1 prompt explicitly instructs models to ``write equations in LaTeX'' (see prompt templates in \Cref{app:prompts}); we therefore include LaTeX typography patterns alongside narrative phrases. \verb|\frac{| alone contributes 23.9M occurrences on DCLM, the single largest F-1 pattern.

\paragraph{Methodology caveats.} Raw occurrence counts reflect token-level prevalence, not document-level coverage. A single equation-heavy document may contain \verb|\frac{| 10+ times while ``step by step'' typically appears once per document. F-1/CoT ratios should therefore be interpreted as upper bounds on document-level prevalence. Nonetheless, the F-1/PoT ratio (121.6$\times$) is robust under any reasonable document normalization, as \verb|\frac{| alone appears in more documents than all PoT T1+T2 patterns combined.

\section{F-1 Pattern Blending Analysis}
\label{app:blending}

This appendix provides the full output n-gram analysis supporting \Cref{sec:blending}. We extract model outputs from Qwen3-30B-Thinking-2507 on RQ1 (F-1+Zero-shot) and RQ2 (F-1+CoT, F-1+PoT) variants, covering all four benchmarks (n=1,456 problems per variant). Pattern counts are case-insensitive string matches; per-output document rates are computed as the fraction of outputs containing at least one occurrence of a given pattern.

\begin{figure}[H]
  \centering
  \includegraphics[width=\columnwidth]{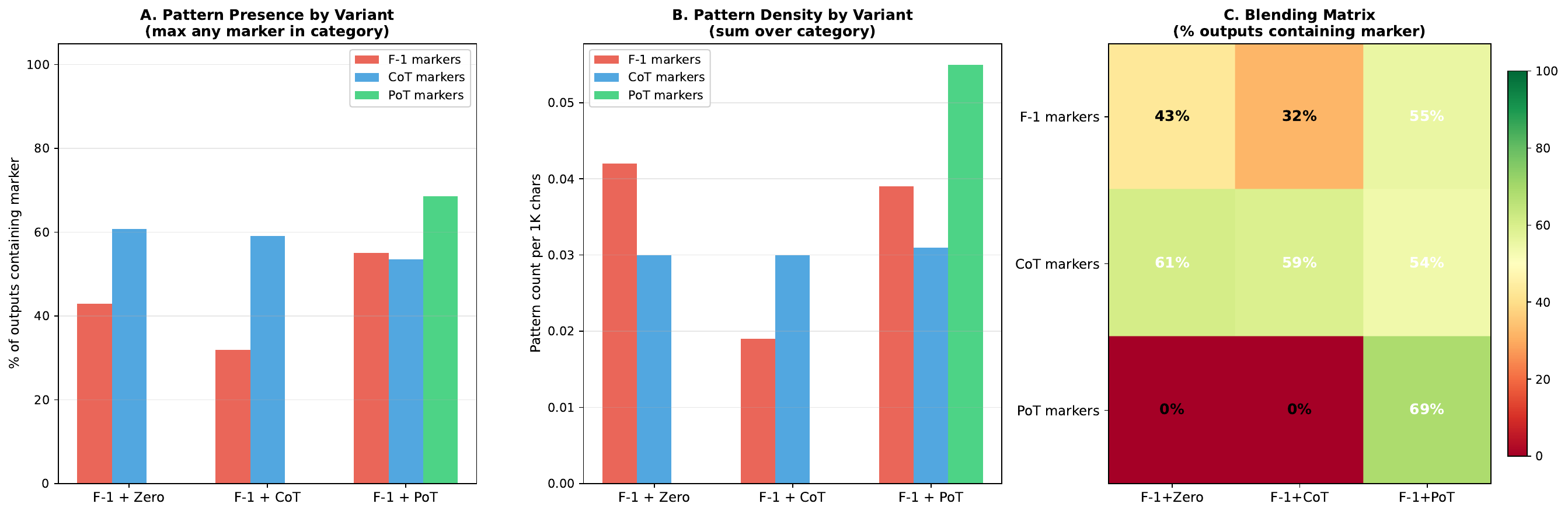}
  \caption{F-1 pattern blending on Qwen3-30B-Thinking ($n{=}1{,}456$ per variant). \textbf{Left:} F-1 markers persist in 43/32/55\% of F-1+Zero/F-1+CoT/F-1+PoT outputs -- F-1 preprocessing survives any solving strategy. \textbf{Center:} Pattern density per 1K chars. \textbf{Right:} Blending matrix -- strategy-specific markers cleanly isolate (PoT: 69\% in F-1+PoT vs 0\% otherwise). ``Step by step'' appears in 54--61\% regardless of variant, confirming it is generic English rather than CoT-specific.}
  \label{fig:blending}
\end{figure}

\paragraph{Why this validates composability.} If F-1 merely replaced CoT/PoT with equation reasoning, we would expect F-1 markers to crowd out strategy-specific markers. Instead, F-1+PoT outputs contain BOTH F-1 equation markers (55\%) AND PoT code markers (69\%), while F-1+CoT contains both F-1 equation markers (32\%) and natural CoT language. The clean isolation of PoT markers at 0\% in non-PoT variants confirms the model respects strategy cues while retaining the F-1 preprocessing phase. This is consistent with F-1 acting as a structural preprocessing prefix that adds equation reasoning without displacing strategy-specific behavior, although a fully causal account would require mechanistic interpretability.

\section{Statistical Tests: Detailed Breakdown}
\label{app:stats}

\Cref{tab:main_results} in the main text reports the overall accuracy results with significance markers. This appendix provides (i) the per-benchmark breakdown of paired tests, (ii) the per-cell McNemar/Wilcoxon results that complement the bolded best-cells in \Cref{tab:main_results}, and (iii) AICrypto-specific bootstrap CIs and a power analysis. The analysis code and full per-model results are available in the accompanying repository.

\begin{table}[!ht]
\centering
\footnotesize
\setlength{\tabcolsep}{2pt}
\begin{tabular}{@{}l ccc@{}}
\toprule
\textbf{Benchmark} & \textbf{vs ZS} & \textbf{vs CoT} & \textbf{vs PoT} \\
 & $\Delta$ / $p$ & $\Delta$ / $p$ & $\Delta$ / $p$ \\
\midrule
Olympiad    & $+$9.71 / $<$.001 & $+$2.32 / $<$.001 & $+$7.37 / $<$.001 \\
IMO-Answer  & $-$0.90 / 0.34 & $-$1.05 / 0.27 & $+$7.95 / $<$.001 \\
\midrule
\textbf{Overall} & $\mathbf{+7.39}$ / $<$\textbf{.001} & $\mathbf{+1.58}$ / $<$\textbf{.001} & $\mathbf{+7.50}$ / $<$\textbf{.001} \\
\bottomrule
\end{tabular}
\caption{Per-benchmark breakdown of paired statistical tests. $\Delta$ in pp; full 95\% CIs in repository. On OlympiadBench, F-1 beats all baselines; on IMO-AnswerBench, F-1 ties CoT ($p{=}0.27$, dominated by difficulty ceiling) but beats PoT by $+$7.95\,pp. Pooling both benchmarks retains significance against CoT (Overall row).}
\label{tab:statistical_tests_full}
\end{table}

\paragraph{Per-cell McNemar / Wilcoxon tests.}
\Cref{tab:per_cell_stats} reports paired tests for every (benchmark, model) cell of \Cref{tab:main_results}. We use McNemar on per-problem correctness for IMO-Bench and OlympiadBench, parse-recovered correctness (round-to-3-decimals equality) for FinanceMath, and Wilcoxon signed-rank on partial-credit scores (0--100, $n{=}18$) for AICrypto.

\begin{table*}[!ht]
\centering
\small
\setlength{\tabcolsep}{6pt}
\begin{tabular}{@{}llrrr@{}}
\toprule
\textbf{Benchmark} & \textbf{Model} & \textbf{$\Delta$ vs ZS / $p$} & \textbf{$\Delta$ vs CoT / $p$} & \textbf{$\Delta$ vs PoT / $p$} \\
\midrule
\multicolumn{5}{@{}l}{\textit{IMO-Bench (n=400, McNemar)}} \\
 & GPT-5          & $-$3.25 / 0.182 & $-$3.25 / 0.182 & $+$15.00 / $<$0.001*** \\
 & Gemini-2.5-Pro & $+$1.25 / 0.630 & $+$0.50 / 0.907 & $+$9.25 / $<$0.001*** \\
 & Qwen3-30B      & $-$0.25 / 1.000 & $+$0.50 / 0.897 & $+$6.75 / 0.003** \\
 & Qwen3-235B     & $-$0.50 / 0.839 & $-$1.50 / 0.286 & $-$8.25 / $<$0.001*** \\
 & DeepSeek-V3.1  & $-$1.75 / 0.500 & $-$1.50 / 0.610 & $+$17.00 / $<$0.001*** \\
\midrule
\multicolumn{5}{@{}l}{\textit{OlympiadBench (n=1{,}438, McNemar)}} \\
 & GPT-5          & $+$2.09 / 0.043*  & $+$2.02 / 0.052 & $+$13.84 / $<$0.001*** \\
 & Gemini-2.5-Pro & $+$25.87 / $<$0.001*** & $+$1.39 / 0.143 & $+$5.63 / $<$0.001*** \\
 & Qwen3-30B      & $+$9.94 / $<$0.001*** & $+$4.10 / $<$0.001*** & $+$5.08 / $<$0.001*** \\
 & Qwen3-235B     & $+$9.79 / $<$0.001*** & $+$5.39 / $<$0.001*** & $+$7.52 / $<$0.001*** \\
 & DeepSeek-V3.1  & $+$0.83 / 0.430 & $-$1.25 / 0.205 & $+$4.80 / $<$0.001*** \\
\midrule
\multicolumn{5}{@{}l}{\textit{FinanceMath (n=200, McNemar on parse-recovered correctness)}} \\
 & GPT-5          & $+$38.00 / $<$0.001*** & $+$21.50 / $<$0.001*** & $+$10.00 / 0.003** \\
 & Gemini-2.5-Pro & $+$24.00 / $<$0.001*** & $+$0.50 / 1.000        & $+$1.00 / 0.845 \\
 & Qwen3-30B      & $+$27.50 / $<$0.001*** & $+$3.50 / 0.230        & $+$2.50 / 0.359 \\
 & Qwen3-235B     & $+$25.50 / $<$0.001*** & $+$26.00 / $<$0.001*** & $+$1.00 / 0.855 \\
 & DeepSeek-V3.1  & $+$16.50 / $<$0.001*** & $+$16.00 / $<$0.001*** & $+$7.00 / 0.088 \\
\midrule
\multicolumn{5}{@{}l}{\textit{AICrypto (n=18 paired, Wilcoxon on 0--100 partial-credit scores)}} \\
 & GPT-5          & $+$7.72 / 0.150 & $-$0.11 / 0.655 & $+$8.91 / 0.128 \\
 & Gemini-2.5-Pro & $+$15.65 / 0.043* & $+$3.25 / 0.180 & $+$4.10 / 0.180 \\
 & Qwen3-30B      & $+$9.75 / 0.063 & $+$18.27 / 0.051 & $+$9.23 / 0.444 \\
 & Qwen3-235B     & $+$8.97 / 0.255 & $+$4.91 / 0.530 & $+$8.36 / 0.289 \\
 & DeepSeek-V3.1  & $+$10.37 / 0.211 & $+$5.92 / 0.495 & $+$3.17 / 0.780 \\
\bottomrule
\end{tabular}
\caption{Per-cell paired tests for \Cref{tab:main_results}. $\Delta$ = F-1 minus baseline (pp). ${*}\,p{<}0.05$, ${**}\,p{<}0.01$, ${***}\,p{<}0.001$, two-sided. F-1 vs ZS: 10/20 cells significant (all 5 FinMath, 4/5 Olympiad, 1/5 AICrypto); F-1 vs CoT: 5/20 (3/5 FinMath: GPT-5, Qwen3-235B, DeepSeek; 2/5 Olympiad: Qwen3 family); F-1 vs PoT: 11/20 (5/5 IMO-Bench, 5/5 Olympiad, 1/5 FinMath, 0/5 AICrypto; note Qwen3-235B IMO is significant in the negative direction at $-8.25$\,pp). AICrypto per-cell tests are underpowered at $n{=}18$; pooled across models, F-1 vs CoT $+6.45$\,pp ($p{=}0.022$); see \Cref{tab:aicrypto_paired_tests}.}
\label{tab:per_cell_stats}
\end{table*}

\paragraph{AICrypto bootstrap CIs and power analysis.}
We defend the $n{=}18$ AICrypto results without removing them: (i) partial-credit per-problem scores (continuous, 0--100) instead of the binary $\geq 80\%$ threshold; (ii) Wilcoxon signed-rank tests with bootstrap CIs (\Cref{tab:aicrypto_paired_tests}); (iii) an explicit power analysis (below).

\begin{table}[!ht]
\centering
\footnotesize
\setlength{\tabcolsep}{3pt}
\resizebox{\columnwidth}{!}{%
\begin{tabular}{@{}lcccccc@{}}
\toprule
& \multicolumn{2}{c}{\textbf{vs ZS}} & \multicolumn{2}{c}{\textbf{vs CoT}} & \multicolumn{2}{c}{\textbf{vs PoT}} \\
\cmidrule(lr){2-3}\cmidrule(lr){4-5}\cmidrule(lr){6-7}
\textbf{Model} & $\Delta$ & $p$ & $\Delta$ & $p$ & $\Delta$ & $p$ \\
\midrule
DeepSeek-V3.1  & $+$10.4 & .211    & $+$5.9  & .495 & $+$3.2  & .780 \\
GPT-5          & $+$7.7  & .150    & $-$0.1  & .655 & $+$8.9  & .128 \\
Gemini-2.5-Pro & $+$15.7 & .043*   & $+$3.3  & .180 & $+$4.1  & .180 \\
Qwen3-235B     & $+$9.0  & .255    & $+$4.9  & .530 & $+$8.4  & .289 \\
Qwen3-30B      & $+$9.8  & .063    & $+$18.3 & .051 & $+$9.2  & .444 \\
\midrule
\textbf{Pooled} & $\mathbf{+10.49}$ & $\mathbf{<.001}$*** & $\mathbf{+6.45}$ & $\mathbf{.022}$* & $\mathbf{+6.75}$ & $\mathbf{.029}$* \\
\bottomrule
\end{tabular}%
}
\caption{AICrypto paired tests on partial-credit scores (per-problem mean, 0--100; per-model $n{=}18$, pooled $n{=}90$). $\Delta$ in pp, F-1 minus baseline. Wilcoxon signed-rank, two-sided. Per-model tests are underpowered at $n{=}18$; pooling 5 models restores significance against all three baselines.}
\label{tab:aicrypto_paired_tests}
\end{table}

\paragraph{Power at $n{=}18$.} For a paired $t$-test at $\alpha{=}0.05$, $80\%$ power, the minimum detectable effect (MDE) given the observed within-pair SD of F-1$-$CoT differences is $19.14$\,pp at the median model and $24.87$\,pp at the noisiest (Cohen's $d_z{=}0.701$). Per-model effects fall below the MDE, which is why per-cell tests are mostly non-significant. Pooling the 90 paired scores across 5 models is the correct level of inference for AICrypto and recovers significance against all three baselines.

\section{Failure-Cause Classification: Multi-Judge Cross-Validation}
\label{app:error_class}

This appendix supplements \S\ref{sec:error_classification}. We sampled $n{=}88$ F-1 failure cases stratified by (benchmark, model): 20 each from FinanceMath, OlympiadBench OE-maths, OE-physics, TP-maths, plus 8 TP-physics (smaller pool); split between GPT-5 and Gemini-2.5-Pro F-1 outputs. Each case was independently classified by three judges of different model lineages -- GPT-5.1 (Azure, low reasoning effort), DeepSeek-V4-Pro thinking, and Claude Opus 4.7 -- using an identical JSON-output prompt. Multi-judge cross-validation defends against single-judge / self-favoring bias \citep{manakul-etal-2023-selfcheckgpt}. Two cases failed to parse on DeepSeek (the other two judges parsed all 88), leaving $n{=}86$ where all three judges produced a valid label.

\begin{table}[H]
\centering
\footnotesize
\setlength{\tabcolsep}{4pt}
\begin{tabular}{@{}lrrrr@{}}
\toprule
\textbf{Cat.} & \textbf{GPT-5.1} & \textbf{DS-V4-Pro} & \textbf{Opus 4.7} & \textbf{Maj.} \\
\midrule
A & 29 (33.0\%) & 22 (25.0\%) & 28 (31.8\%) & 22 (25.6\%) \\
B &  2 (2.3\%)  &  0 (0.0\%)  &  3 (3.4\%)  &  1 (1.2\%) \\
C & 57 (64.8\%) & 64 (72.7\%) & 57 (64.8\%) & 63 (73.3\%) \\
\midrule
\textit{Unp.} & 0 & 2 & 0 & --- \\
\bottomrule
\end{tabular}
\caption{Per-judge marginal classifications of $n{=}88$ F-1 failures. \textbf{A} = wrong-eq $\rightarrow$ wrong-ans (over-reliance); \textbf{B} = wrong-eq $\rightarrow$ recovered (resilience); \textbf{C} = right-eq $\rightarrow$ exec-err (mechanical). All three judges converge: most failures (64--73\%) preserve a correct Phase-1 equation; 25--33\% are over-reliance; self-correction (B) is rare (0--3\%). Majority-vote (right column, $n{=}86$ where all three judges parsed) ties on category C dominance.}
\label{tab:error_class}
\end{table}

\paragraph{Inter-rater agreement.} Pairwise Cohen's $\kappa$: GPT-5.1 vs DeepSeek-V4-Pro = $0.522$ (moderate); GPT-5.1 vs Claude Opus 4.7 = $0.306$ (fair); DeepSeek vs Claude = $0.385$ (fair). Three-way Fleiss' $\kappa{=}0.403$ (moderate, $n{=}86$); 52 of 86 cases (60.5\% [95\% CI: 49.9, 70.1]) had unanimous agreement across all three judges. While individual case-level agreement is fair-to-moderate, the marginal distributions converge tightly, supporting the high-level claim that Phase-1 formalization is reliable on $\sim$73\% of failure cases regardless of which judge is used.

\paragraph{What this defends.} Single-judge classification (especially when the judge belongs to the same family as the model under test) risks self-favoring bias. By using three judges of independent lineages -- and reporting the worst-case pairwise $\kappa$ alongside the majority-vote rate -- we make the dominant finding (C ${\gg}$ A ${\gg}$ B) robust to any single judge's idiosyncrasies.

\section{Phase-1 Verification Probe}
\label{app:verify_probe}

This appendix supplements the Limitations discussion on Phase-1 over-reliance (\S\ref{sec:limitations}, item 1). The failure-cause analysis (\Cref{app:error_class}) finds that 25.6\% of F-1 failures are wrong-equation $\rightarrow$ wrong-answer, with only 1.2\% self-correcting. A natural mitigation is to add an explicit Phase-1 self-verification step. We test whether such a step recovers a measurable fraction of these failures \emph{within a single call}.

\paragraph{Variant.} We define F-1+Verify by inserting a Phase 1.5 instruction between Phase 1 and Phase 2:

\begin{promptbox}[title=F-1+Verify Phase 1.5 instruction]
\small\ttfamily
Phase 1.5 -- Verify:\\
Re-read the problem and check whether the equations correctly represent every given, the target, and any constraints. If you find a mismatch, sign error, missing variable, or wrong governing equation, revise the equations below before continuing. If correct, state ``Equations verified.'' and proceed.
\end{promptbox}

System prompt and Phase 1/Phase 2 instructions are identical to FP/F-1+ZS on FinanceMath. F-1+Verify therefore differs from FP \emph{only} in the addition of Phase 1.5.

\paragraph{Setup.} FinanceMath validation ($n{=}200$, rule-based scoring, $\epsilon{=}10^{-6}$). Model: GPT-5 (high reasoning) -- the cell with the largest F-1 lift over CoT in \Cref{tab:main_results} and the clearest deployment of the Phase-1 over-reliance failure mode. Both conditions use the same dataset order and decoding settings.

\begin{table}[H]
\centering
\small
\setlength{\tabcolsep}{4pt}
\begin{tabular}{@{}lrrrrr@{}}
\toprule
\textbf{Method} & \textbf{Acc.} & \textbf{Net flips} & \textbf{$\Delta$} & \textbf{Tokens} & \textbf{$\Delta$tok} \\
\midrule
F-1 (FP)         & 64.00 & ---  &        & 2{,}243 &  --- \\
F-1+Verify       & 64.50 & $+$1 & $+$0.50 & 2{,}890 & $+$647 \\
\bottomrule
\end{tabular}
\caption{F-1+Verify vs F-1 on FinanceMath~$\times$~GPT-5 ($n{=}200$, paired). \textbf{Acc.} = accuracy~(\%). \textbf{Net flips} = (only-F-1+Verify-correct) $-$ (only-F-1-correct) = $9-8$. \textbf{Tokens} = mean completion tokens per problem. McNemar exact two-sided $p{=}1.0$.}
\label{tab:verify_probe}
\end{table}

\paragraph{Results.} F-1+Verify recovers 9 problems that F-1 misses but breaks 8 that F-1 solves correctly, for a net gain of 1 problem out of 200 ($+0.50$\,pp, McNemar exact $p{=}1.0$). The probe also incurs a substantial token cost: $+647$ mean completion tokens per problem ($+28.8\%$) and $+115$ mean prompt tokens. Crucially, in 200/200 outputs the model concludes Phase 1.5 with ``Equations verified.'' or a textually equivalent acknowledgement; no output revises its Phase 1 equations during Phase 1.5. The behavior the verification step is designed to elicit -- self-detection and correction of a wrong governing equation -- never fires.

\paragraph{Interpretation.} The probe converges with the failure-cause finding (\Cref{app:error_class}) that self-correction is rare ($\text{B}{=}1.2\%$ in the multi-judge analysis, $0\%$ here): once the model has committed to a governing equation in Phase 1, explicit single-call self-verification does not surface the error. The over-reliance is structural, not addressable by adding a verification turn within the same generation. Multi-call backtracking \citep{yao2023tree,liu2023xot} -- where Phase 1 outputs are evaluated by a separate call before solving -- remains the more promising mitigation, at the cost of leaving the single-call regime that F-1 targets.

\paragraph{Scope.} This probe runs on a single (benchmark, model) cell with the largest expected gain and the cleanest failure-mode signal. Negative findings here do not preclude the possibility that a different verification prompt formulation, a different benchmark, or a smaller / non-reasoning model could yield different behavior. The result establishes that the most direct single-call mitigation is not effective for the cell that should most benefit from it.

\section{F-1 vs Formulate-and-Solve: Per-Problem Complementarity}
\label{app:fs_complementarity}

This appendix supplements the F-1 vs F\&S discussion in \S\ref{sec:related}. We re-implemented Formulate-and-Solve following \citet{kao2024formulate} -- a single-call prompt that asks the model to (i) formulate the problem as a system of equations over named variables, then (ii) solve algebraically -- and ran it on FinanceMath ($n{=}200$) and OlympiadBench Math (OE\_TO\_maths\_en\_COMP, $n{=}674$) with two of our paper's models: GPT-5 (proprietary, frontier) and Qwen3-30B-A3B-Thinking-2507 (open, mid-tier). \Cref{tab:fs_complementarity} cross-tabulates per-problem correctness in all four (model, benchmark) cells.

\begin{table}[!ht]
\centering
\footnotesize
\setlength{\tabcolsep}{3pt}
\resizebox{\columnwidth}{!}{%
\begin{tabular}{@{}llrrrrrr@{}}
\toprule
\textbf{Bench} & \textbf{Model} & \textbf{N} & \textbf{F-1} & \textbf{F\&S} & \textbf{F1$\setminus$FS} & \textbf{FS$\setminus$F1} & \textbf{Union} \\
\midrule
FinMath & GPT-5    & 200 & 64.00 & 64.00 & 5.5 & 5.5 & 69.5 \\
FinMath & Qwen30B  & 200 & \textbf{56.00} & 51.50 & 9.5 & 5.0 & 61.0 \\
OB Math & GPT-5    & 674 & 86.35 & \textbf{87.09} & 3.4 & 4.2 & 90.5 \\
OB Math & Qwen30B  & 674 & 78.64 & 78.64 & 7.9 & 7.9 & 86.5 \\
\bottomrule
\end{tabular}%
}
\caption{F-1 vs Formulate-and-Solve \citep{kao2024formulate} per-problem cross-tabulation. \textbf{F-1} / \textbf{F\&S} columns: aggregate accuracy (\%). \textbf{F1$\setminus$FS} / \textbf{FS$\setminus$F1}: \% of problems solved only by F-1 / only by F\&S (set-difference). \textbf{Union} = (both$+$F1-only$+$FS-only)/$N$. Aggregate accuracies are close (within $\pm$5 pp) but the methods solve substantially different problem subsets: 7.6--15.7\% disjoint across cells. Combining both methods exceeds either alone by $+$4--8\,pp, confirming F-1's adaptive Phase 2 is orthogonal to F\&S's fixed algebraic-solve pipeline. Best per-cell aggregate accuracy in \textbf{bold}.}
\label{tab:fs_complementarity}
\end{table}

\paragraph{Reading the table.} On FinanceMath, F-1 wins on Qwen3-30B by $+$4.5 pp and ties GPT-5; on OB Math, F\&S wins on GPT-5 by $+$0.74 pp and ties Qwen3-30B. None of the per-cell aggregate gaps are large. \emph{However}, the disjoint-success columns reveal that even when accuracies tie, the two methods are solving different problems: e.g., on Qwen3-30B OB Math both methods score $530/674$ but $53$ problems are solved only by F-1 and $53$ only by F\&S -- a 15.7\% disjoint set. This is direct evidence that F-1's contribution is qualitatively different from F\&S's, not a marginal accuracy improvement on the same problem distribution. The union (both methods combined as an oracle) reaches $86.5$--$90.5$\% on OB Math and $61$--$70$\% on FinanceMath, indicating substantial unrealized headroom from method ensembling.

\paragraph{Why aggregate accuracies tie despite disjoint success sets.} The per-cell ties (e.g., Qwen3-30B OB Math both at $530/674$) are not coincidence: each method solves a comparable number of problems, but the \emph{identities} of those problems differ. F-1 tends to win on multi-step problems requiring strategy switching mid-solution -- where its Phase 2 menu (CoT/PoT/Direct) lets the model defer the solving-mode choice until after equation formalization. F\&S, with its fixed algebraic-solve pipeline, tends to win on problems where a clean equation system is the bottleneck and downstream solving is mechanical. The two failure modes are largely orthogonal, which is why aggregate counts converge while individual problems diverge.

\paragraph{Implication for method ensembling.} The union accuracies suggest a simple downstream ensemble: run both F-1 and F\&S, accept the answer when they agree, defer to a verifier when they disagree. On Qwen3-30B OB Math, this would bound test-time accuracy at $86.50\%$ (the union) -- a $+7.86$ pp lift over either method alone. We do not pursue ensembling here because our focus is single-call methods under fixed compute, but it is a promising direction for follow-up work targeting applied-math benchmarks where governing equations exist but solution structure varies.

\section{BBH Negative Control: Per-Strategy Details}
\label{app:bbh_negctrl}

This appendix supplements \S\ref{sec:negative_control}. We applied F-1 verbatim (Phase-1 formalization + Phase-2 strategy menu) to BBH word\_sorting \citep{suzgun-etal-2023-challenging} -- a task that requires alphabetically sorting a list of 5--10 English words and contains no equations or quantitative structure. We ran three strategies: \textbf{Zero-Shot}, \textbf{CoT} (``think step by step''), and \textbf{F-1} (full prefix). Models tested: GPT-5 (Azure), Gemini-2.5-Pro, and Qwen3-30B-A3B-Thinking-2507 served via vLLM. Each cell evaluates 250 problems.

\begin{table}[!ht]
\centering
\footnotesize
\setlength{\tabcolsep}{4pt}
\begin{tabular}{@{}lrrr@{}}
\toprule
\textbf{Model} & \textbf{ZS} & \textbf{CoT} & \textbf{F-1} \\
\midrule
GPT-5             & 98.8 (247) & 96.0 (240) & \textbf{46.8 (117)} \\
Gemini-2.5-Pro    & 87.6 (219) & 98.4 (246) & \textbf{56.0 (140)} \\
Qwen3-30B-Th.     & 94.0 (235) & 93.2 (233) & 88.4 (221) \\
\bottomrule
\end{tabular}
\caption{BBH word\_sorting full results across 3 models and 3 strategies (each cell: accuracy\%~(correct/250)). F-1 regression cells in bold for GPT-5 and Gemini ($-$49.2 and $-$42.4\,pp vs CoT). Qwen3-30B-Thinking regresses only $-$4.8\,pp, reflecting that thinking-mode models partially adapt away from inappropriate prompt instructions during their internal reasoning trace.}
\label{tab:bbh_negctrl_full}
\end{table}

\paragraph{Why F-1 fails on word\_sorting.} The Phase-1 directive (``identify governing equations, write them in LaTeX'') has no semantic referent for a list of words. Inspection of GPT-5 F-1 outputs shows the model attempting to (i) cast sorting as an algebraic permutation problem with formal index variables, (ii) write LaTeX expressions like $\sigma : \{w_1,\ldots,w_n\} \to \{w_{\sigma(1)},\ldots,w_{\sigma(n)}\}$, and (iii) ``solve'' for the permutation -- a route that produces consistent answer-format errors and occasional sorting mistakes. The instruction-following models obey the formalization directive literally; the thinking-mode model partially overrides it during its internal trace, which is why Qwen3-30B-Thinking regresses substantially less.

\paragraph{Implication.} This cleanly delineates F-1's deployment envelope: F-1 is a targeted intervention for problems with identifiable governing equations (applied math, physics, finance, engineering), \emph{not} a generic prompting upgrade. We recommend practitioners restrict F-1 to domains where Phase 1's instruction has a meaningful semantic referent.

\section{Document-Level Pattern Normalization}
\label{app:doc_norm}

This appendix supplements the robustness check in \S\ref{sec:infinigram}. The token-level infini-gram counts (\Cref{app:infini-gram}, \Cref{fig:eq-vs-code}) report total occurrences of method-associated phrases. A reviewer concern is that LaTeX typography commands (e.g., \verb|\frac{|) can appear multiple times per equation-heavy document, so token-level prevalence may overstate document-level coverage. We address this by sampling random occurrences per pattern and counting the number of unique source documents.

\paragraph{Sampling protocol.} For each pattern, we drew $K{=}100$ random occurrence locations via the infini-gram \texttt{find} endpoint and resolved each to a source document ID via the \texttt{get\_doc\_by\_rank} endpoint. The unique-document count $U$ over $K$ samples gives an estimate of average occurrences per document, $\widehat{occ/doc} \approx K/U$. Combined with the token total $T$, the estimated number of distinct documents containing the pattern is $\widehat{D} = T \cdot U / K$. We restrict the sampling to DCLM-baseline (the curated, highest-quality corpus) and to a representative subset of T1+T2 patterns (10 CoT, 10 PoT, 21 F-1 patterns covering both narrative phrases and LaTeX environments).

\begin{table}[!ht]
\centering
\small
\setlength{\tabcolsep}{4pt}
\begin{tabular}{@{}lrrr@{}}
\toprule
\textbf{Method} & \textbf{Tokens} & \textbf{Est.\ Docs} & \textbf{Tok/Doc} \\
\midrule
CoT  & 18{,}829{,}256 & 18{,}826{,}446 & 1.00 \\
PoT  &  1{,}183{,}071 &  1{,}179{,}270 & 1.00 \\
F-1  & 10{,}755{,}424 & 10{,}739{,}381 & 1.00 \\
\bottomrule
\end{tabular}
\caption{Document-level normalization of infini-gram pattern counts on DCLM-baseline. Estimated occurrences per document are $\approx$1.0 across all three methods, indicating that token-level ratios are not artifactually inflated by repeat-occurrence concentration in a small set of documents.}
\label{tab:doc_norm}
\end{table}

\paragraph{Result.} \Cref{tab:doc_norm} reports the aggregate token counts, estimated unique document counts, and tokens-per-document for each method on DCLM. The tokens-per-document ratio is $\approx$1.0 for all three methods (CoT 1.0001, PoT 1.0032, F-1 1.0015), so each pattern occurrence is, on average, in its own distinct document. This holds even for high-volume LaTeX patterns: \verb|\begin{equation}| (451K tokens, 451K docs), \verb|\begin{align}| (467K tokens, 462K docs). The two patterns where occ/doc deviates from 1.0 (``key governing equations'' at 8.88, ``equations in latex'' at 1.67) have very small token totals (71 and 886 respectively) and contribute negligibly to the aggregate.

\paragraph{Implication.} The token-level F-1/CoT and F-1/PoT ratios reported in \S\ref{sec:infinigram} closely track their document-level counterparts -- the document-coverage caveat does not materially change the ranking. F-1 patterns are genuinely (not artifactually) rarer than CoT in pretraining; the gain mechanism is structural ordering of equation-first prompting at inference, not a corpus-statistical artifact.

\FloatBarrier

\section{AI Usage Statement}

LLMs were used solely to assist with writing tasks, such as proofreading, grammatical corrections, and \LaTeX{} formatting. All research ideas, primary content, and initial drafts were created by the authors, and LLMs did not generate the core manuscript material. LLMs were also used to support code generation and debugging during evaluation and analysis. All resulting content are carefully reviewed and validated by the authors.

\end{document}